\documentclass[10pt,twocolumn,letterpaper]{article}
\usepackage[pagenumbers]{cvpr} 
\usepackage{multirow}
\usepackage{colortbl}
\usepackage{float}
\usepackage{tabularx}
\usepackage{pifont}
\usepackage{booktabs}
\usepackage{amsmath,amssymb}
\usepackage{algorithm}
\usepackage{algpseudocode}
\definecolor{expTableRow}{RGB}{241,246,251}
\definecolor{expTableSecond}{RGB}{36,104,170}
\definecolor{expTableCite}{RGB}{36,104,170}
\definecolor{expTableGain}{HTML}{009901}
\newcommand{\gain}[1]{\textcolor{expTableGain}{#1}}

\usepackage{amsthm}

\newtheorem{proposition}{Proposition}
\newtheorem{lemma}{Lemma}
\AtBeginEnvironment{theorem}{\par\nopagebreak}
\AtBeginEnvironment{proposition}{\par\nopagebreak}
\AtBeginEnvironment{lemma}{\par\nopagebreak}
\AtBeginEnvironment{proof}{\par\nopagebreak}

\definecolor{cvprblue}{rgb}{0.21,0.49,0.74}
\usepackage[pagebackref,breaklinks,colorlinks,allcolors=cvprblue]{hyperref}

\def\confName{CVPR}
\def\confYear{2027}

\title{R\textsuperscript{3}Sieve: Suppressing Unconditional Information from Attention to Answers \\ for High-Resolution VQA}

\author{%
  \textbf{Yifan Cao}\textsuperscript{1}\thanks{Equal contribution.},
  \textbf{Yawen Zhang}\textsuperscript{2}\footnotemark[1],
  \textbf{Xiaocui Yang}\textsuperscript{1}\thanks{Corresponding author.},
  \textbf{Faxian Wan}\textsuperscript{1},
  \textbf{Shi Feng}\textsuperscript{1}\\
  \textbf{Daling Wang}\textsuperscript{1},
  \textbf{Yifei Zhang}\textsuperscript{1},
  \textbf{Shangshu Yu}\textsuperscript{1},
  \textbf{Xianbing Zhao}\textsuperscript{3}\\[3pt]
  {\normalsize\textsuperscript{1}School of Computer Science and Engineering,
  Northeastern University, Shenyang 110819, China}\\
  {\normalsize\textsuperscript{2}School of Biomedical Engineering \&
  State Key Laboratory of Advanced Medical Materials and Devices,}\\
  {\normalsize ShanghaiTech University, Shanghai 201210, China}\\
  {\normalsize\textsuperscript{3}Jiangnan University, Wuxi, China}\\[3pt]
  {\small\texttt{\{caoyifan,wanfaxian\}@mails.neu.edu.cn}}\\
  {\small\texttt{\{yangxiaocui,fengshi,wangdaling,zhangyifei,shangshuyu\}@cse.neu.edu.cn}}\\
  {\small\texttt{zhangyw2025@shanghaitech.edu.cn},
  \texttt{zhaoxianbing\_hitsz@163.com}}%
}

\begin{document}
\maketitle
\begin{abstract}
    High-resolution visual question answering (VQA) requires models to identify question-relevant small objects in high-resolution images and select the correct answer based on what they recognize, demanding fine-grained perception. Previous methods focus on obtaining better visual evidence (image crops), but obtaining it alone is insufficient, using it effectively is equally crucial. Visual evidence should therefore be utilized as effectively as possible throughout inference. We argue that small-target evidence is obscured by \textbf{unconditional information} that remains observable in subsequent processing even after the condition on which it should depend is removed. Such information hinders the effective use of small-target evidence. We propose R$^{3}$Sieve (\textbf{R}esidual-\textbf{R}egion-\textbf{R}esponse Sieve), a training-free framework. We divide inference into three stages---localization, cropping, and answering---and introduce distinct counterfactual references to suppress interference at each stage. During localization, the Shared-Subspace Canceller (SSC) uses a generic image description to construct a counterfactual reference and obtain question-relevant attention maps. During cropping, the Unique-Contribution Spotter (UCS) constructs counterfactual references from visual and non-visual features to identify distinctive, complementary crop regions. During answering, the Content-Aligned Voter (CAV) constructs an image-free branch to calibrate answer responses. Experiments on three high-resolution VQA benchmarks show that R$^{3}$Sieve achieves the highest overall accuracy among compared methods across diverse settings, improving base models by up to 21.5 percentage points.
\end{abstract}

\begin{figure}[t]
  \centering
  \includegraphics[width=\linewidth]{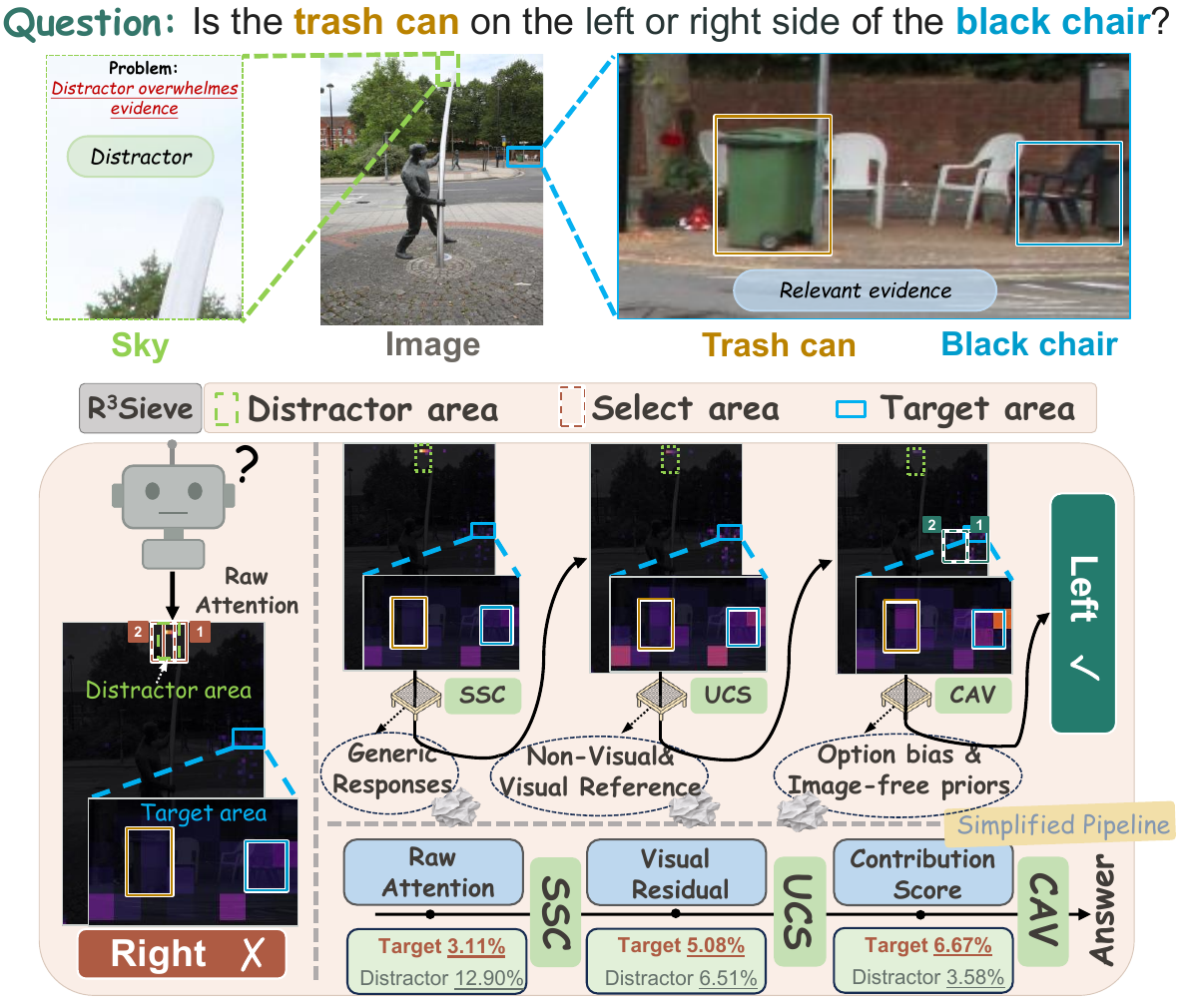}
  \caption{
Unconditional information can disrupt high-resolution VQA. Here, raw attention to question-irrelevant sky impairs localization, causing errors. R$^{3}$Sieve suppresses interference during localization (SSC), cropping (UCS), and answering (CAV), progressively reducing distractor responses and strengthening target responses (bottom), yielding the correct answer.
}
  \label{fig:intro}
\end{figure}    
\begin{figure*}[t]
    \centering
    \includegraphics[width=\textwidth]{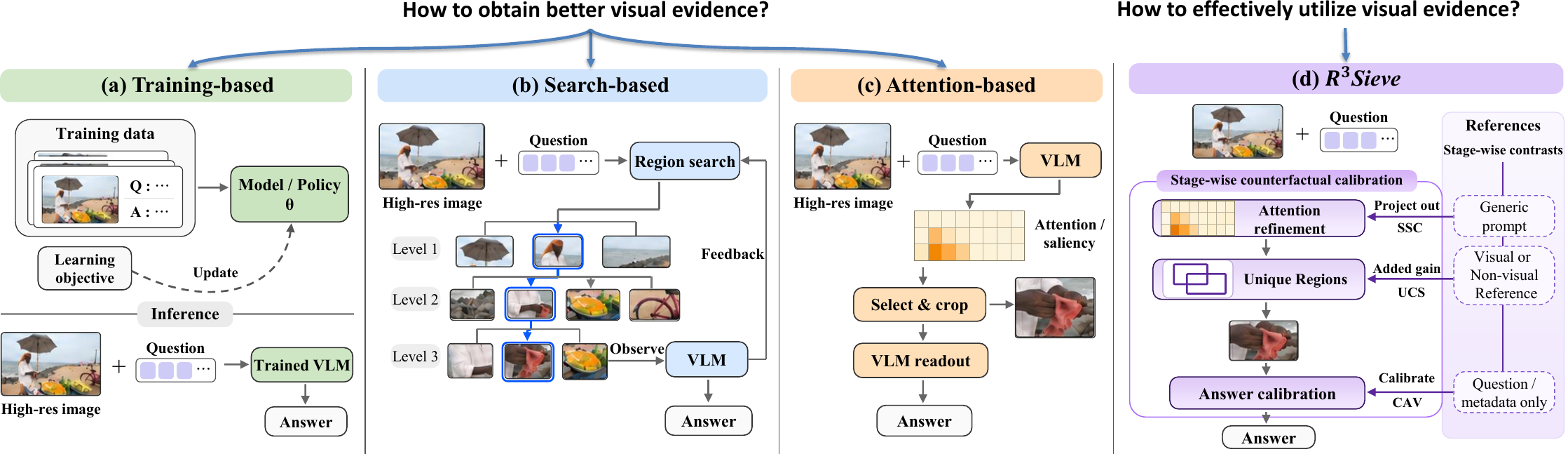}
    \caption{
    Comparison of high-resolution VQA approaches. (a) Training-based methods acquire visual evidence. (b) Search-based methods locate relevant regions iteratively. (c) Attention-based methods guide cropping with attention signals. (d) Our R$^{3}$Sieve uses all-stage counterfactual filtering to suppress unconditional information during localization, cropping, and answering, improving evidence utilization.
    }
    \label{fig:method_taxonomy}
\end{figure*}

\section{Introduction}
\label{sec:intro}

Multimodal large language models (MLLMs)~\cite{llava,bai2025qwen3vltechnicalreport,internvl3,internvl35} have advanced visual understanding and reasoning, yet high-resolution images remain challenging for fine-grained perception~\cite{vstar,vicrop}. Question-relevant evidence, such as object attributes and spatial relations, often occupies small regions amid irrelevant content, requiring models to locate, preserve, and use visual details. 
Existing approaches improve evidence acquisition and processing in different ways. Training-based methods (Fig.~\ref{fig:method_taxonomy}(a)) include V$^\ast$~\cite{vstar}, which combines guided search with trained localization, PixelEyes~\cite{pixeleye} for mask-guided visual search, and ZwZ~\cite{zwz} and ReVuE~\cite{revue} for distilling regional observations or evidence-aware reasoning. Training-free approaches can be grouped by primary guidance mechanism. Search-guided methods (Fig.~\ref{fig:method_taxonomy}(b)), including BVS~\cite{bvs}, EviSpec~\cite{evispec}, and the zero-shot configuration of SPARC~\cite{pmlr-v306-avogaro26a}, use search optimization, evidence specifications, or explicit visual search. Attention-guided methods (Fig.~\ref{fig:method_taxonomy}(c)), including ViCrop~\cite{vicrop} and HiDe~\cite{hide}, select or process regions using attention-derived signals. However, identifying relevant regions does not ensure that their evidence is preserved during cropping or effectively used during answer generation.

Recent studies show that attention may cover queried targets even when models answer incorrectly~\cite{seeingnotbelieving,seeingrightsayingwrong}. In our experiments, all annotated targets satisfy our attention coverage criterion in at least one decoding layer for 80.0\%--93.9\% of incorrect responses (Supp.~\ref{app:attention_errors}). This raises a question: \emph{what prevents MLLMs from using evidence they already attend to?} We argue that small-target evidence is obscured by \emph{unconditional information}, the part of a response that persists when the condition it should depend on is removed. Such interference appears differently across inference. During localization, attention responds to salient backgrounds, such as the sky in Fig.~\ref{fig:intro}, and generic structures independently of the question~\cite{damro}. During cropping, the relevant condition is the spatial location itself: responses that remain explainable after removing a location are redundant. Such responses consume the limited crop budget and displace weaker but complementary cues, particularly for spatial relations involving two objects, such as the trash can and black chair in Fig.~\ref{fig:intro}. During answering, option-order bias~\cite{optionorder,pride} and language priors~\cite{vcd} favor answers unsupported by the image. Because these forms arise at different stages, suppressing interference only during localization, as in some attention-guided methods, leaves other sources of distortion unaddressed.

To fully utilize visual evidence, we propose \textbf{R$^{3}$Sieve} (Residual-Region-Response Sieve) (Fig.~\ref{fig:method_taxonomy}(d)), a training-free framework that removes stage-specific unconditional information using counterfactual references. Across localization, cropping, and answering, R$^{3}$Sieve follows a unified counterfactual residual principle, using responses from the same MLLM to construct references capturing what remains without the relevant condition and preserve unexplained contributions. At the localization stage, the Shared-Subspace Canceller (SSC) builds a reference subspace from generic-description responses and projects it out of question-conditioned attention, yielding question-relevant visual residuals. For cropping, the Unique-Contribution Spotter (UCS) uses visual and non-visual responses to score each location by its unique residual contribution. Greedy Coverage Selection (GCS) selects complementary multi-scale regions by their marginal gains in coverage and detail. Finally, at the answering stage, we introduce the Content-Aligned Voter (CAV), which pools scores across option permutations~\cite{pride} and subtracts image-free predictions~\cite{calibrate} to calibrate responses against option-order and language priors. Thus, the three R's represent the visual \emph{Residual}, complementary \emph{Regions}, and calibrated \emph{Response}.

As illustrated in Fig.~\ref{fig:intro}, successive filtering increases the target-to-distractor ratio from raw attention to residual and contribution scores, changing an incorrect answer into the correct one. Across MLLMs and high-resolution benchmarks, R$^{3}$Sieve achieves state-of-the-art accuracy among the compared methods while filtering interference throughout inference.
Our main contributions are as follows:

\begin{itemize}
    \item We introduce \emph{unconditional information} and experimentally demonstrate its presence in localization, cropping, and answering, extending interference suppression beyond a single stage.
 
    \item We propose \textbf{R$^{3}$Sieve}, a training-free framework that removes unconditional information at all three stages with one counterfactual residual operation.
 
    \item Across three high-resolution benchmarks and four Qwen3-VL/InternVL3.5 models, R$^{3}$Sieve achieves the highest overall accuracy among compared methods, improving base models by up to 21.5 points. It also runs faster than compared training-free methods on HRBench-4K/8K and remains robust on general VQA benchmarks.
\end{itemize}

\section{Related Work}
\label{sec:related}

\paragraph{Multimodal Large Language Models.}
Multimodal pretraining and instruction tuning have advanced
visual understanding and instruction following in
MLLMs~\cite{blip2,instructblip,llava}.
High-resolution perception further requires preserving local
details within a limited visual-token budget.
Qwen3-VL~\cite{bai2025qwen3vltechnicalreport} integrates
multi-level visual features, while
InternVL~\cite{internvl3,internvl35},
Monkey~\cite{monkey}, and LLaVA-UHD~\cite{llavauhd}
preserve high-resolution information through tiling, patching,
or adaptive slicing.
Training-based methods further strengthen local perception.
PixelEyes~\cite{pixeleye} links reasoning with localization,
Learning to Zoom Efficiently~\cite{curriculum} learns zooming
through contrastive training, and
Zooming without Zooming~\cite{zwz} and ReVuE~\cite{revue}
distill local observations from teachers or visual agents.
These methods improve the representation and acquisition of
local details, providing a foundation for high-resolution
visual question answering.

\paragraph{High-Resolution Visual Perception.}

Training-free high-resolution VQA methods mainly follow search-based or attention-based approaches (Fig.~\ref{fig:method_taxonomy}(b,c)). Search-based approaches include hierarchical search in ZoomEye~\cite{zoomeye}, Monte Carlo tree search in DyFo~\cite{dyfo}, Bayesian optimization in BVS~\cite{bvs}, and visual search or evidence retrieval in zero-shot SPARC~\cite{pmlr-v306-avogaro26a}, RAP~\cite{rap}, and EviSpec~\cite{evispec}. Attention-based methods, including ViCrop~\cite{vicrop} and HiDe~\cite{hide}, use attention signals to guide region selection.
Yet relevant crops may be redundant and miss complementary objects, while attention coverage alone does not ensure correct answers~\cite{seeingnotbelieving,seeingrightsayingwrong}. Related studies investigate attention suppression, logit contrast, and evidence selection~\cite{rtwi}. At answer readout, prior methods contrast distorted-image or language-only predictions~\cite{vcd,lcd}, estimate preferences from content-free inputs~\cite{calibrate}, or correct option-position bias through permutations~\cite{pride}. Yet these approaches do not jointly address interference across inference stages.
In contrast, R$^{3}$Sieve applies all-stage counterfactual filtering (Fig.~\ref{fig:method_taxonomy}(d)). SSC removes generic attention patterns, UCS scores unique regional contributions against other responses, GCS selects complementary crops, and CAV calibrates answers using option permutations and image-free predictions. Together, these stages improve evidence localization, selection, and utilization.

\begin{figure*}[t]
  \centering
  \includegraphics[width=\linewidth]{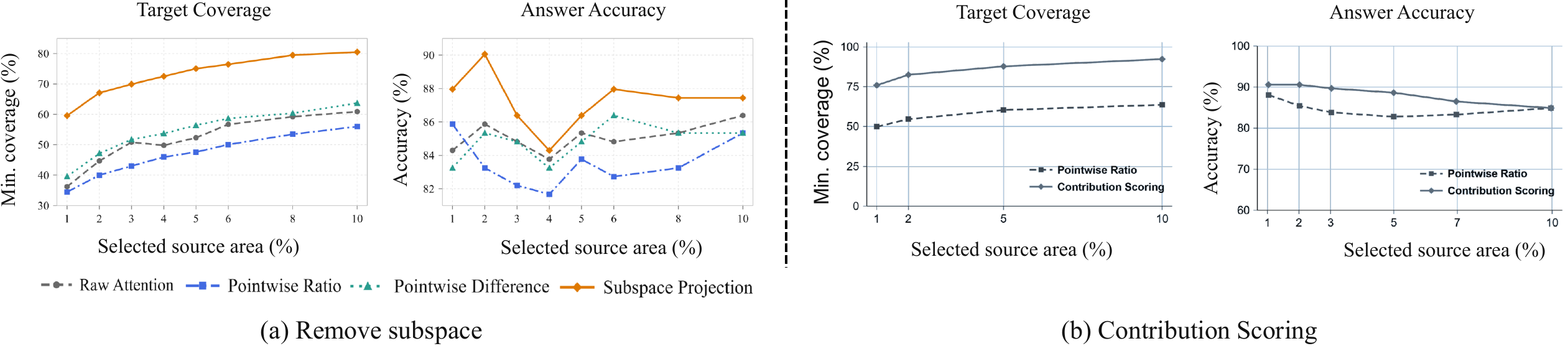}
  \caption{Unconditional information in localization and cropping. (a) Subspace Projection versus pointwise corrections on minimum target coverage and answer accuracy. (b) Contribution Scoring versus Pointwise Ratio under the same source-area budgets.}
  \label{fig:subspace_an}
\end{figure*}

\section{Unconditional Information}
\label{sec:interference}
We first formalize unconditional information and then locate it at each stage of inference. Let $x$ denote the input to a stage, such as an image--question pair, and let $f(x)$ be a signal that should depend on a condition $z$ of $x$. A counterfactual input $\tilde{x}$ removes $z$ or changes only a factor that the signal should ignore. The unconditional component of $f(x)$ is the part that the counterfactual response explains, and removing it leaves the residual that the evidence adds,
\begin{equation}
\label{eq:unconditional}
f^{\mathrm{u}}(x)=\Pi\bigl[f(x)\mid f(\tilde{x})\bigr],
\quad
f^{\mathrm{r}}(x)=f(x)-f^{\mathrm{u}}(x),
\end{equation}
where the superscripts $\mathrm{u}$ and $\mathrm{r}$ denote the unconditional component and residual, respectively, and $\Pi[\cdot\mid\cdot]$ takes the part of the first argument explained by the second. The following experiments identify interference from unconditional information at each stage and quantify the gains from suppressing it with counterfactual references, motivating our method. All analyses use Qwen3-VL-4B on V$^*$Bench, and the option-order analysis also covers HRBench-4K and HRBench-8K.
 
\subsection{Unconditional Information in Localization}
\label{sec:visual_interference}
During target localization, the question is the condition. As observed by ViCrop~\cite{vicrop}, attention can favor salient backgrounds and common structures independent of the question. To remove these responses, we pair the same image with the question and the generic prompt “Describe the image.” to form factual and counterfactual inputs. Normalizing their visual attention and taking element-wise square roots gives $H$ and $H_0$, with one row per prompt token. Let $B_0$ be a row-orthonormal basis of the row space of $H_0$. Inspired by All-but-the-Top~\cite{allbuttop}, Subspace Projection removes the generic-response component from $H$,
\begin{equation}
\label{eq:residual}
R=H-(HB_0^\top)B_0,
\qquad
d_j=\lVert R_{:,j}\rVert_2^2,
\end{equation}
where $(HB_0^\top)B_0$ is the unconditional component, $R$ is the visual residual, and $d_j$ scores token $j$.

We compare Subspace Projection with Raw Attention, Pointwise Difference adapted from HiDe~\cite{hide}, and Pointwise Ratio adapted from ViCrop~\cite{vicrop}. Let $\bar a$ and $\bar a_0$ denote token-averaged question and generic attention. Pointwise Difference subtracts $\bar a_0$ from each question-token map, whereas Pointwise Ratio uses $\bar a/\bar a_0$ (Supp.~\ref{app:background-processing}). Unlike these single-map corrections, Subspace Projection removes the entire generic-response subspace. Under identical source-area and token budgets, it achieves the highest minimum target coverage and answer accuracy across all evaluated budgets (Fig.~\ref{fig:subspace_an}(a), definitions in Supp.~\ref{app:selection_metrics}), suggesting that generic interference spans multiple spatial patterns.

\begin{figure}[t]
\centering
\includegraphics[width=\linewidth]{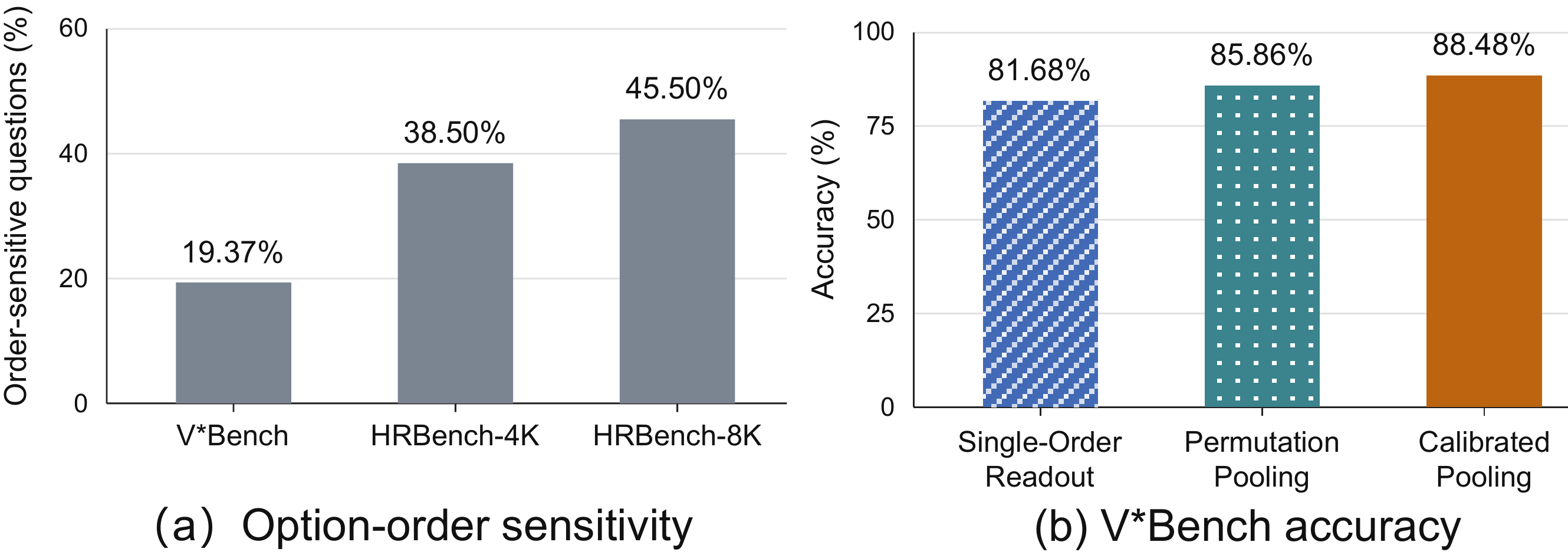}
\caption{Unconditional information in answer generation. (a) Option-order sensitivity across three benchmarks. (b) Accuracy gains from Permutation Pooling and Calibrated Pooling on V$^*$Bench~\cite{vstar}.}
\label{fig:contribution}
\end{figure}

\subsection{Unconditional Information in Region Cropping}
\label{sec:complementary}
In region cropping, the condition is the location. Responses that other locations can reproduce are redundant and consume the limited crop budget. To suppress this redundancy, we propose Contribution Scoring. We use the $j$-th visual residual $r_j$ from $R$ as the signal and the remaining columns of $[R\,N]$ as the counterfactual reference, where $N$ contains normalized square-root attention to nonvisual tokens. Along direction $w$, the signal and reference energies are $(w^\top r_j)^2$ and $w^\top G_{-j}w$, with $G_{-j}=RR^\top+NN^\top-r_jr_j^\top$. Contribution Scoring maximizes their log ratio,
\begin{equation}
\label{eq:contribution}
\begin{aligned}
\log q_j
&=\max_{w\neq0}\Bigl[
\log(w^\top r_j)^2
-\log\bigl(w^\top G_{-j}w\bigr)
\Bigr],\\
\phi_j
&=r_j^\top\bigl(RR^\top+NN^\top\bigr)^{-1}r_j
=\frac{q_j}{1+q_j}.
\end{aligned}
\end{equation}
Here, $\log(w^\top G_{-j}w)$ represents the reference energy, while $\phi_j$ measures a location's unique contribution. If $m\ge2$ locations have identical residuals, then $\phi_j\le1/m$ regardless of their magnitude. Thus, repeated responses are discounted, while nonvisual responses prevent weak but isolated visual patterns from receiving inflated scores.

Under identical source-area and token budgets, Contribution Scoring improves minimum target coverage and matches or exceeds Pointwise Ratio in answer accuracy (Fig.~\ref{fig:subspace_an}(b)). This supports discounting redundant responses to reserve crop budgets for complementary evidence.

\subsection{Unconditional Information in Answering}
\label{sec:question_prior}
During answer generation, predictions should depend on the image and option content rather than option position or language priors. However, reversing option order while keeping the image, question, and answer contents fixed changes 19.37\%, 38.50\%, and 45.50\% of predicted answers on V*Bench, HRBench-4K, and HRBench-8K, respectively (Fig.~\ref{fig:contribution}(a)). We interpret this sensitivity as interference from unconditional information associated with option position. To suppress these interference, let $\ell_\pi(c)$ denote the log-probability of answer content $c$ under option order $\pi$, with $K$ orders. Reordering alters only option positions, yielding counterfactual references with the same question and option content as the original input. Pooling across orders yields $M_{\mathrm v}(c)$, suppressing position-dependent preferences. The pooled image-free score $M_{\mathrm q}(c)$ captures preferences that persist without visual evidence. We subtract both counterfactual components,
\begin{equation}
\label{eq:calibrated}
\begin{aligned}
M_{\mathrm v}(c)&=\frac{1}{a}\log\frac{1}{K}\sum_{\pi'}e^{a\ell_{\pi'}(c)},\\
s(c)&=\ell_\pi(c)-\bigl[\ell_\pi(c)-M_{\mathrm v}(c)\bigr]-\beta M_{\mathrm q}(c),
\end{aligned}
\end{equation}
where $a\neq0$ controls pooling and $\beta>0$ weights the image-free reference. Single-Order Readout, Permutation Pooling, and Calibrated Pooling use $\ell_\pi(c)$, $M_{\mathrm v}(c)$, and $s(c)$, respectively.

\begin{figure*}[t]
  \centering
  \includegraphics[width=\linewidth]{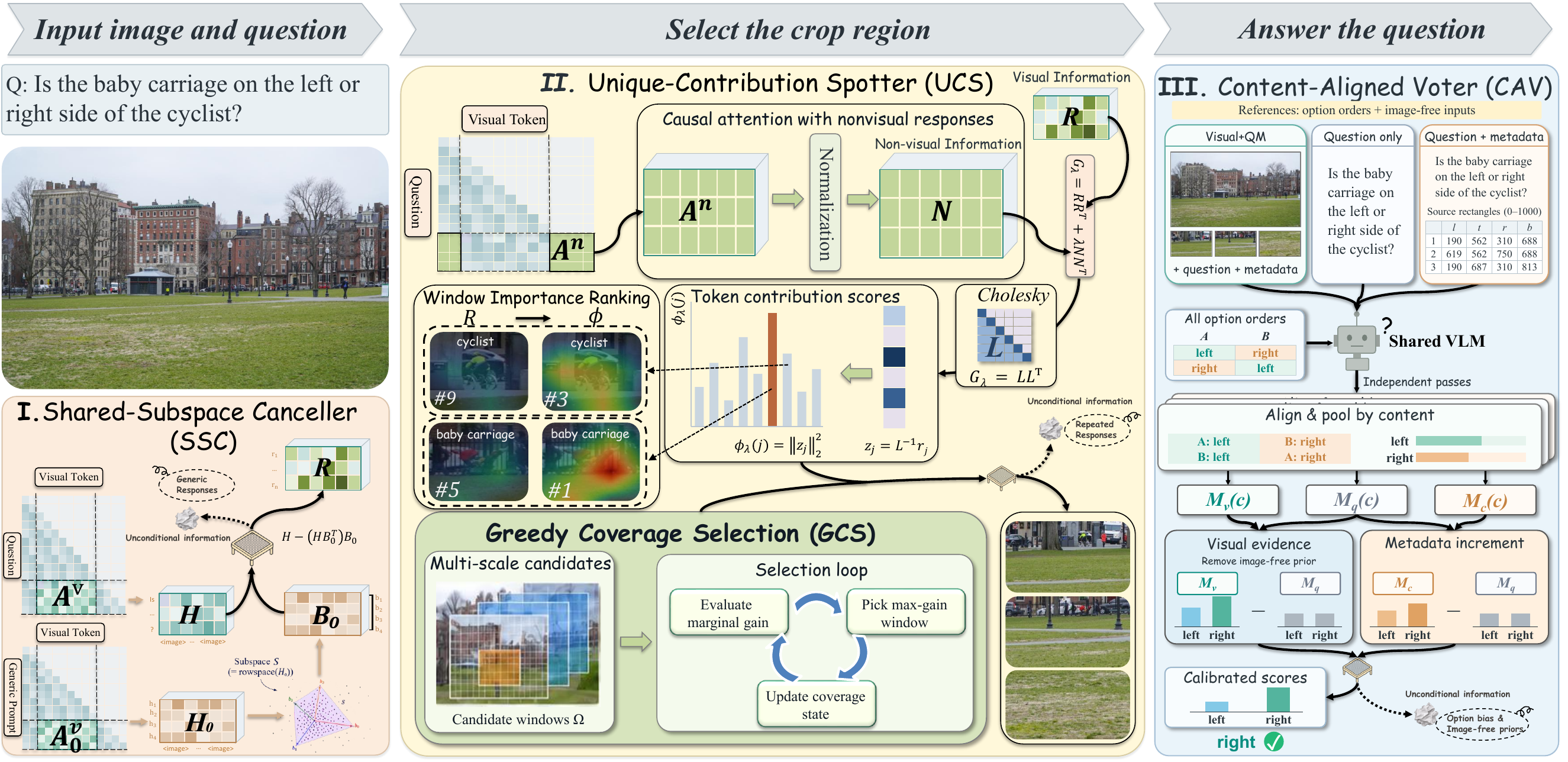}
  \caption{Overview of R$^{3}$Sieve. Each module removes the unconditional information of one stage with a counterfactual reference. SSC removes the attention that a generic prompt explains, UCS discounts responses that other locations repeat and selects complementary crops through GCS, and CAV pools answer scores over option orders by content and subtracts the image-free preference.}
  \label{fig:main_figure}
\end{figure*}

On V$^*$Bench, Permutation Pooling increases accuracy from 81.68\% to 85.86\%, and Calibrated Pooling further improves it to 88.48\% (Fig.~\ref{fig:contribution}(b)). Together with localization and cropping, these results show that stage-specific counterfactual references suppress distinct forms of unconditional information under the shared residual principle of Eq.~\eqref{eq:unconditional}.

\section{Method}
\label{sec:method}
Building on the analysis in Sec.~\ref{sec:interference},
we propose R$^{3}$Sieve to suppress interference from
unconditional information throughout the VQA pipeline,
including target localization, region cropping, and answer
generation, using corresponding counterfactual references
(Fig.~\ref{fig:main_figure}). Given an image $I$, a question $q$, and candidate answers $\mathcal C$, the Shared-Subspace Canceller (SSC) extracts the visual residual from a preview, the Unique-Contribution Spotter (UCS) selects complementary regions from this residual, and the Content-Aligned Voter (CAV) scores candidates using the full image and selected crops to produce a calibrated response. In this section, we describe the algorithm in detail, with the corresponding pseudocode provided in Supp.~\ref{app:algorithm}.

\subsection{Shared-Subspace Canceller (SSC)}
\label{sec:shared_subspace}
As shown in Fig.~\ref{fig:main_figure} (I), SSC removes the attention that a generic prompt explains. We run the preview twice, with the question and with the generic prompt, average the attention over heads at decoder layer $d$, and keep the text-query rows. UCS and window selection also run on this preview to save computation. Let $A^{\mathrm v}\in\mathbb R^{T\times V}$ and $A^{\mathrm n}\in\mathbb R^{T\times U}$ be the attention from the $T$ question tokens to the $V$ visual and $U$ nonvisual keys. We normalize both by the total visual attention $m_t=\sum_jA^{\mathrm v}_{tj}$ of each query, which keeps their relative strength, and take element-wise square roots,
\begin{equation}
\label{eq:visual_attention_transform}
H_{tj}=\sqrt{\frac{A^{\mathrm v}_{tj}}{\max(m_t,\epsilon)}},
\qquad
N_{tk}=\sqrt{\frac{A^{\mathrm n}_{tk}}{\max(m_t,\epsilon)}},
\end{equation}
with $\epsilon=10^{-12}$ for numerical stability. The same transform gives $H_0$ for the generic prompt, and $B_0$ is its orthonormal row-space basis after negligible eigenvalues are discarded (details in Supp.~\ref{app:construction_subspace}). SSC computes $R$ via Eq.~\eqref{eq:residual}, where $(HB_0^\top)B_0$ is the unconditional component.

\begin{proposition}[Orthogonal residual]
\label{prop:subspace}
$R$ is the closest matrix to $H$ in Frobenius norm whose rows are orthogonal to $B_0$, and $\lVert R\rVert_F^2=\lVert H\rVert_F^2-\lVert HB_0^\top\rVert_F^2$.
\end{proposition}
 
The projection thus discards exactly the energy that the generic responses explain (details in Supp.~\ref{app:proof_subspace}), so UCS starts from attention that carries no generic pattern.

\subsection{Unique-Contribution Spotter (UCS)}
\label{sec:unique_contribution}
UCS discounts responses repeated at other locations, turning the remaining evidence into crops (Fig.~\ref{fig:main_figure} (II)).
\paragraph{Contribution scoring.}
UCS weights the nonvisual reference of Eq.~\eqref{eq:contribution} by $\lambda>0$,
\begin{equation}
\label{eq:visual_token_contribution}
G_\lambda=RR^\top+\lambda NN^\top,
\qquad
\phi_\lambda(j)=r_j^\top G_\lambda^{-1}r_j,
\end{equation}
which reduces to $\phi_j$ for $\lambda=1$. Since $G_\lambda=X_\lambda X_\lambda^\top$ with $X_\lambda=[R,\sqrt{\lambda}N]$, $\phi_\lambda(j)$ is the statistical leverage score of column $j$ in $X_\lambda$~\cite{mahoney}, measuring its contribution beyond what the other visual and nonvisual columns provide. A larger $\lambda$ further lowers the scores of locations that only linear independence supports, and all scores follow from the Cholesky factor $L$ of $G_\lambda$ as $\phi_\lambda(j)=\lVert(L^{-1}R)_{:,j}\rVert_2^2$. Two standard properties of leverage scores, proved in Supp.~\ref{app:proof_repeat}, make the discount of repeated responses explicit.

\begin{proposition}[Discount of repeated responses]
\label{prop:repeat}
If $m$ other locations repeat $r_j$ as $r_{k_i}=\alpha_ir_j$, then $\phi_\lambda(j)\le1/(1+\sum_{i=1}^m\alpha_i^2)$.
\end{proposition}

A response repeated with equal strength at $m$ other locations thus scores at most $1/(m+1)$. GCS complements this token-level discount by accounting for coverage overlap (details in Supp.~Table~\ref{tab:scoring_selection}).

\paragraph{Greedy Coverage Selection (GCS).}
GCS turns the scores into at most three crops that observe high-scoring locations at fine resolution. The set $\Omega$ contains the Cartesian products of intervals that slide along each preview-grid axis with lengths $\{1,1/2,1/4,1/8\}$ of the axis and a half-length stride, excluding the full image. A window $u$ with $|u|$ visual tokens has the detail gain $\eta(u)=(V/|u|)^\rho-1$, where $\rho>0$ favors smaller windows. Each location keeps the largest gain among the selected windows that contain it, and a selected set $S$ is scored by
\begin{equation}
\label{eq:coverage_objective}
\begin{aligned}
g_S(j)&=\max\bigl(\{\eta(u)\mid u\in S,\ j\in u\}\cup\{0\}\bigr),\\
F(S)&=\textstyle\sum_{j=1}^{V}\phi_\lambda(j)\,g_S(j).
\end{aligned}
\end{equation}
Starting from $S=\varnothing$, GCS repeatedly adds the window with the largest marginal gain
\begin{equation}
\label{eq:window_marginal_gain}
\Delta(u\mid S)=\sum_{j\in u}\phi_\lambda(j)\bigl[\eta(u)-g_S(j)\bigr]_+,
\end{equation}
Here, $[x]_+=\max\{x,0\}$ and $\Delta(u\mid S)=F(S\cup\{u\})-F(S)$. GCS discounts redundant coverage, much as Eq.~\eqref{eq:contribution} discounts repeated responses, and stops after three windows or when no positive gain remains.

\begin{proposition}[Approximation guarantee of GCS]
\label{prop:gcs}
$F$ is monotone and submodular, so greedy selection of $k$ windows attains at least $1-(1-1/k)^k\geq1-1/e$ of the optimal coverage.
\end{proposition}
For three windows, the factor is $19/27$, following the classical result~\cite{nemhauser} (details in Supp.~\ref{app:gcs-guarantee}). The selected windows are cropped from the original image and answered together with the full image.

\subsection{Content-Aligned Voter (CAV)}
\label{sec:content_aligned_voter}
As shown in Fig.~\ref{fig:main_figure} (III) CAV addresses the two answer preferences found in Sec.~\ref{sec:question_prior}. Within each branch, we normalize option-letter logits at temperature $T_v$ for the visual branch or $T_t$ for both text-only branches, then map log-probabilities
to answer contents to align scores across orders. Thus scores from different orders refer to the same content. With $\ell_\pi(c)$ the log-probability of content $c$ under order $\pi$, we pool $K$ orders with mellowmax~\cite{mellowmax}, for $a\ne0$,
\begin{equation}
\label{eq:permutation_score_aggregation}
M^{(a)}(c)=\frac{1}{a}\log\Bigl[
\frac{1}{K}\sum_{\pi=1}^{K}
\exp\bigl(a\,\ell_\pi(c)\bigr)
\Bigr].
\end{equation}
For analysis, define the mean extension $M^{(0)}(c)=K^{-1}\sum_{\pi=1}^{K}\ell_\pi(c)$. Positive $a$ favors high-scoring orders, while negative $a$ favors low-scoring ones.

\begin{proposition}[Order invariance and ranking preservation]
\label{prop:order}
With other branch inputs fixed, pooling all $n!$ orders of $n$ distinct contents is order-invariant for any real $a$. If normalized log-probabilities satisfy $\ell_\pi(c)=g(c)+b(p_\pi(c))+\kappa_\pi$, where $g$ is an order-independent content score, $p_\pi(c)$ is the position of $c$, $b$ is positional bias, and $\kappa_\pi$ is a normalization offset shared across contents, then $M^{(a)}$ preserves $g$'s ranking, including ties.
\end{proposition}

The image-free preference is estimated by two text-only branches. The visual branch receives the full image, crops, crop metadata, question, and options; the question-only branch receives the question and options, and the metadata-only branch additionally receives crop metadata (details in Supp.~\ref{metadata_template}). Pooling with $a_{\mathrm v}$ for the visual branch and $a_{\mathrm t}$ for both text branches yields
$M_{\mathrm{v}}^{(a_{\mathrm{v}})}(c)$,
$M_{\mathrm{q}}^{(a_{\mathrm{t}})}(c)$, and
$M_{\mathrm{c}}^{(a_{\mathrm{t}})}(c)$,
and the calibrated score is
\begin{equation}
\label{eq:language_prior_calibration}
\begin{aligned}
s(c)={}&M_{\mathrm{v}}^{(a_{\mathrm{v}})}(c)
-\beta M_{\mathrm{q}}^{(a_{\mathrm{t}})}(c)\\
&+\gamma\bigl[
M_{\mathrm{c}}^{(a_{\mathrm{t}})}(c)
-M_{\mathrm{q}}^{(a_{\mathrm{t}})}(c)
\bigr],
\end{aligned}
\end{equation}
where $\beta$ weights the image-free reference in Eq.~\eqref{eq:calibrated} and $\gamma$ weights the increment due to crop metadata. CAV answers with $c^\star=\arg\max_{c\in\mathcal C}s(c)$. Supp.~\ref{app:cav_diagnostics} evaluates the effects of pooling exponents, metadata correction, and temperature scaling. All orders in a branch share the common-prefix KV cache, so the visual input is encoded once and each order evaluates only its text suffix.
\begin{table*}[t]
\centering
\caption{Comparison on high-resolution fine-grained perception benchmarks (accuracy, \%). Best/second-best results are \textbf{bold}/\textcolor{expTableSecond}{blue}. TF: training-free. $^\dagger$: indicates results taken from the original papers due to unavailable model weights or excessive evaluation time.}  
\label{tab:exp_main}

\begingroup
\normalsize
\setlength{\tabcolsep}{3pt}
\renewcommand{\arraystretch}{0.85}

\setlength{\aboverulesep}{0.2ex}
\setlength{\belowrulesep}{0.3ex}

\hypersetup{citecolor=expTableCite}

\begin{tabular}{@{\hspace{5pt}}clc*{9}{r}@{\hspace{5pt}}}
\toprule
\multirow{2}{*}{Backbone} & \multirow{2}{*}{Method} & \multirow{2}{*}{TF} & \multicolumn{3}{c}{V$^*$Bench} & \multicolumn{3}{c}{HRBench-4K} & \multicolumn{3}{c}{HRBench-8K}\\
\cmidrule(lr){4-6}\cmidrule(lr){7-9}\cmidrule(lr){10-12}
& & & Attr. & Spa. & All & FSP & FCP & All & FSP & FCP & All\\
\midrule
\multirow{6}{*}{Qwen3-VL-4B} & w/o reasoning & $\checkmark$ & 85.2 & 76.3 & 81.7 & 87.3 & 64.3 & 75.8 & 86.3 & 62.0 & 74.1 \\
 & BVS~\cite{bvs}\,{\small\itshape (ICML'26)} & $\checkmark$ & \textcolor{expTableSecond}{97.4} & 82.9 & 91.6 & \textcolor{expTableSecond}{92.0} & 71.5 & \textcolor{expTableSecond}{81.8} & \textcolor{expTableSecond}{93.3} & 63.5 & 78.4\\
 & SPARC $^\dagger$~\cite{pmlr-v306-avogaro26a}\,{\small\itshape (ICML'26)}& $\checkmark$ & -- & -- & 87.0 & -- & -- & 80.4 & -- & -- & 78.6\\
 & ZwZ ~\cite{zwz}\,{\small\itshape (ICML'26)} & $\times$ & 94.8 & \textcolor{expTableSecond}{89.5} & \textcolor{expTableSecond}{92.7} & 89.0 &\textcolor{expTableSecond}{71.8} & 80.4 & 86.8 & \textcolor{expTableSecond}{70.8} & \textcolor{expTableSecond}{78.8}\\
\rowcolor{expTableRow}
 & \textbf{R$^{3}$Sieve (ours)} & $\checkmark$ & \textbf{98.3} & \textbf{90.8} & \textbf{95.3} & \textbf{94.0} & \textbf{74.0} & \textbf{84.0} & \textbf{94.0} & \textbf{72.0} & \textbf{83.0}\\
 \rowcolor{expTableRow}
 & \gain{$\Delta$ vs.\ Qwen3-VL-4B} &
 & \gain{+13.1} & \gain{+14.5} & \gain{+13.6}
 & \gain{+6.7}  & \gain{+9.7}  & \gain{+8.2}
 & \gain{+7.7}  & \gain{+10.0} & \gain{+8.9}\\
\midrule
\multirow{7}{*}{Qwen3-VL-8B} & w/o reasoning & $\checkmark$ & 84.3 & 80.3 & 82.7 & 91.0 & 64.5 & 77.8 & 85.3 & 62.3 & 73.8 \\
 & BVS~\cite{bvs}\,{\small\itshape (ICML'26)} & $\checkmark$ & 93.9 & \textcolor{expTableSecond}{89.5} & 92.1 & \textcolor{expTableSecond}{96.5} & \textcolor{expTableSecond}{71.8} & 84.1 & 93.3 & 64.3 & 78.8\\
 & HiDe~\cite{hide}\,{\small\itshape (ICML'26)} & $\checkmark$ & 91.3 & 85.5 & 89.0 & 95.0 & 70.3 & 82.6 & \textcolor{expTableSecond}{95.0} & \textcolor{expTableSecond}{67.8} & 81.4\\
 & EviSpec $^\dagger$~\cite{evispec}\,{\small\itshape (arXiv 2026.09)} & $\checkmark$ & \textcolor{expTableSecond}{94.8} & 86.8 & 91.6 & 94.8 & 70.3 & 82.5 & 94.0 & 66.8 & 80.4\\
 & PixelEyes $^\dagger$~\cite{pixeleye}\,{\small\itshape (arXiv 2026.06)} & $\times$ & -- & -- & \textcolor{expTableSecond}{94.2} & -- & -- & \textcolor{expTableSecond}{85.0} & -- & -- & \textcolor{expTableSecond}{83.2}\\
\rowcolor{expTableRow}
 & \textbf{R$^{3}$Sieve (ours)} & $\checkmark$ & \textbf{96.5} & \textbf{92.1} & \textbf{94.8} & \textbf{100} & \textbf{73.0} & \textbf{86.5} & \textbf{98.0} & \textbf{72.0} & \textbf{85.0}\\
 \rowcolor{expTableRow}
 & \gain{$\Delta$ vs.\ Qwen3-VL-8B} &
 & \gain{+12.2} & \gain{+11.8} & \gain{+12.1}
 & \gain{+9.0}  & \gain{+8.5}  & \gain{+8.7}
 & \gain{+12.7} & \gain{+9.7}  & \gain{+11.2}\\
\midrule
\multirow{5}{*}{InternVL3.5-4B} & w/o reasoning & $\checkmark$ & 71.3 & 75.0 & 72.8 & 76.0 & 53.8 & 64.9 & 63.8 & 48.3 & 56.0 \\
 & BVS~\cite{bvs}\,{\small\itshape (ICML'26)} & $\checkmark$ & \textcolor{expTableSecond}{92.2} & \textcolor{expTableSecond}{84.2} & \textcolor{expTableSecond}{89.0} & \textcolor{expTableSecond}{90.5} & \textcolor{expTableSecond}{54.3} & 72.4 & \textcolor{expTableSecond}{88.8} & \textcolor{expTableSecond}{52.8} & \textcolor{expTableSecond}{70.8}\\
 & ReVuE $^\dagger$~\cite{revue}\,{\small\itshape (arXiv 2026.09)} & $\times$ & -- & -- & 73.3 & -- & -- & \textcolor{expTableSecond}{73.2} & -- & -- & 67.4\\
\rowcolor{expTableRow}
 & \textbf{R$^{3}$Sieve (ours)} & $\checkmark$ & \textbf{94.8} & \textbf{86.8} & \textbf{91.6} & \textbf{91.0} & \textbf{66.0} & \textbf{78.5} & \textbf{90.0} & \textbf{65.0} & \textbf{77.5}\\
 \rowcolor{expTableRow}
 & \gain{$\Delta$ vs.\ InternVL3.5-4B} &
 & \gain{+23.5} & \gain{+11.8} & \gain{+18.8}
 & \gain{+15.0} & \gain{+12.2} & \gain{+13.6}
 & \gain{+26.2} & \gain{+16.7} & \gain{+21.5}\\
\midrule
\multirow{5}{*}{InternVL3.5-8B} & w/o reasoning & $\checkmark$ & 73.9 & 72.4 & 73.3 & 74.5 & \textcolor{expTableSecond}{57.0} & 65.8 & 69.3 & 54.0 & 61.6 \\
 & BVS ~\cite{bvs}\,{\small\itshape (ICML'26)} & $\checkmark$ & \textcolor{expTableSecond}{93.9} & \textbf{88.2} & \textcolor{expTableSecond}{91.6} & \textcolor{expTableSecond}{91.0} & 55.8 & \textcolor{expTableSecond}{73.4} & \textcolor{expTableSecond}{89.0} & \textcolor{expTableSecond}{56.5} & 72.8\\
 & Curriculum $^\dagger$~\cite{curriculum}\,{\small\itshape (EMNLP'26)} & $\times$ & -- & -- & 81.2 & -- & -- & 72.8 & -- & -- & \textcolor{expTableSecond}{72.9}\\
\rowcolor{expTableRow}
 & \textbf{R$^{3}$Sieve (ours)} & $\checkmark$ & \textbf{97.4} & \textcolor{expTableSecond}{84.2} & \textbf{92.1} & \textbf{95.0} & \textbf{74.0} & \textbf{84.5} & \textbf{91.0} & \textbf{74.0} & \textbf{82.5}\\
 \rowcolor{expTableRow}
 & \gain{$\Delta$ vs.\ InternVL3.5-8B} &
 & \gain{+23.5} & \gain{+11.8} & \gain{+18.8}
 & \gain{+20.5} & \gain{+17.0} & \gain{+18.7}
 & \gain{+21.7} & \gain{+20.0} & \gain{+20.9}\\
\bottomrule
\end{tabular}
\endgroup
\end{table*}

\section{Experiments}
\label{sec:experiments}

\subsection{Experimental Settings}
\label{sec:exp_setup}

\paragraph{Benchmarks and metrics.}
We evaluate accuracy on V$^*$Bench~\cite{vstar} for attribute recognition (Attr.) and spatial reasoning (Spa.), and HRBench-4K/8K~\cite{hrbench} for Fine-grained Single-instance Perception (FSP) and Fine-grained Cross-instance Perception (FCP). All denotes overall.
 
\paragraph{Backbones and baselines.}
We apply R$^{3}$Sieve to Qwen3-VL~\cite{bai2025qwen3vltechnicalreport} and InternVL3.5~\cite{internvl35}, using their 4B and 8B variants. Baselines include training-free BVS~\cite{bvs}, HiDe~\cite{hide}, EviSpec~\cite{evispec}, and SPARC~\cite{pmlr-v306-avogaro26a}, and training-based PixelEyes~\cite{pixeleye}, ZwZ~\cite{zwz}, ReVuE~\cite{revue}, and Curriculum~\cite{curriculum}.
 
\paragraph{Implementation details.}
R$^{3}$Sieve localizes evidence using decoder-layer $d$ attention on a preview, selects up to three crops, and answers using the crops and full image. Main experiments use NVIDIA RTX A6000 GPUs and all others use NVIDIA A100-80GB GPUs. Hyperparameters and evaluation protocols are detailed in Supp.~\ref{app:experiment_details}, with sensitivity analysis in Supp.~\ref{app:sensitivity}. Tables~\ref{tab:exp_main}--\ref{tab:exp_transfer} use the same visual token budget (Supp.~\ref{app:visual_tokens}).

\begin{table}[t]
\centering
\small
\caption{Component ablation on V$^*$Bench (\%). Bold: best.}
\label{tab:component_ablation}
\setlength{\tabcolsep}{1.6pt}
\renewcommand{\arraystretch}{0.9}
\setlength{\aboverulesep}{0.2ex}
\setlength{\belowrulesep}{0.3ex}
\begin{tabularx}{\linewidth}{@{}l*{3}{>{\centering\arraybackslash}X}@{}}
\toprule
Components & Attr. & Spa. & All\\
\midrule
\rowcolor{black!10}
\textbf{R$^{3}$Sieve} & \textbf{98.26} & \textbf{90.79} & \textbf{95.29}\\
w/o SSC & 97.39 & 88.16 & 93.72\\
w/o UCS & \textbf{98.26} & 86.84 & 93.72\\
w/o CAV & \textbf{98.26} & 85.53 & 93.19\\
w/o (SSC + UCS) & 97.39 & 85.53 & 92.67\\
w/o (SSC + CAV) & 97.39 & 85.53 & 92.67\\
w/o (UCS + CAV) & \textbf{98.26} & 82.89 & 92.15\\
w/o All (Base) & 93.91 & 76.32 & 86.91\\
\bottomrule
\end{tabularx}
\end{table}

\subsection{Main Results}

Table~\ref{tab:exp_main} compares R$^{3}$Sieve with training-free and training-based methods on four backbones. Without additional training, our method achieves the highest overall accuracy in all 12 backbone--benchmark settings. The gains extend to both attribute recognition and spatial or cross-instance reasoning, suggesting that improving evidence utilization benefits more than a single question type. For every backbone, the gain on HRBench-8K exceeds that on HRBench-4K. As image resolution increases, small targets occupy a smaller fraction of the visual input, making generic and redundant responses more likely to compete with useful evidence. SSC suppresses generic attention, UCS preserves complementary regions, and CAV reduces answer preferences that need not follow the image. This stage-wise design provides a plausible explanation for the larger gains on harder high-resolution inputs. InternVL3.5 benefits most, improving by 13.6--21.5 points and reducing its gap to the same-size Qwen3-VL models from 8.9--18.1 points to at most 5.5 points. Together with the high attention-coverage rates for incorrect answers (Supp.~\ref{app:attention_errors}), these results support the importance of filtering interference after relevant evidence has already been attended to.

\subsection{Ablation Study}
\label{sec:exp_ablation}

\begin{table}[t]
    \centering
    \caption{Transferability of CAV.  Best results are bold.}
    \label{tab:cav_transfer}
    \small
    \setlength{\tabcolsep}{1.6pt}
    \renewcommand{\arraystretch}{0.9}
    \setlength{\aboverulesep}{0.2ex}
    \setlength{\belowrulesep}{0.3ex}
    \begin{tabularx}{\linewidth}{@{}l*{3}{>{\centering\arraybackslash}X}
    }
        \toprule
        Method
        & V$^*$Bench
        & HRBench-4K
        & HRBench-8K \\
        \midrule
        HiDe~\cite{hide}
        & 91.62 & 79.38 & 78.38 \\
        
        HiDe + CAV
        & 92.67 & 82.50 & 82.00 \\
        
        BVS~\cite{bvs}
        & 91.62 & 81.75 & 78.38 \\
        
        BVS + CAV
        & 92.15 & 81.50 & 82.00 \\
        
        \rowcolor{black!10}
        \textbf{R$^{3}$Sieve}
        & \textbf{95.29}
        & \textbf{84.00}
        & \textbf{83.00} \\
        \bottomrule
    \end{tabularx}
\end{table}

\paragraph{Component ablation study.}
\label{sec:component_ablation}
To evaluate the contributions of the three stages, we compare different combinations of SSC, UCS, and CAV using Qwen3-VL-4B (Table~\ref{tab:component_ablation}). The common base ranks candidate windows independently by raw attention and selects three crops.

SSC achieves 98.26\% attribute accuracy, suggesting removing question-independent attention helps locate small attribute-bearing objects. UCS reaches 85.53\% spatial accuracy by prioritizing complementary regions, helping preserve evidence for both objects in a spatial relation. Combining SSC and UCS retains both advantages, indicating that localization and cropping address distinct interference sources. CAV further improves spatial accuracy by 2.63--9.21 points across configurations. Since spatial questions often involve opposite relations such as left and right, option-order bias and language priors can affect predictions even when relevant evidence is available. CAV mitigates these biases through calibration.
The full model achieves 95.29\% overall accuracy, outperforming the base by 8.38 points and SSC + UCS by 2.10 points. This demonstrates that answer-stage interference suppression remains beneficial after improving evidence localization and selection.

\paragraph{Transferability of CAV.}
\label{sec:transferability_ablation}

We apply CAV to HiDe~\cite{hide} and BVS~\cite{bvs} without changing their visual inputs or token budgets (Table~\ref{tab:cav_transfer}). It improves five of six settings, suggesting that answer-stage biases persist after region selection. The exception is BVS on HRBench-4K, dropping from 81.75\% to 81.50\%. Representative cases suggest that calibration can overturn narrow correct-answer margins in attribute and counting questions (Supp.~\ref{app_bvs_cav_cases}). Thus, CAV complements but cannot replace reliable evidence selection .
Further ablations separate scoring from selection and evaluate CAV under frozen settings (Supp.~\ref{app:additional_ablations}).

\subsection{Additional Experiments}
\label{sec:exp_additional}

\begin{table}[t]
\centering
\caption{Inference time (second/question), shown as mean (SD).}
\label{tab:efficiency}
 
\begingroup
\small
\setlength{\tabcolsep}{1.6pt}
\renewcommand{\arraystretch}{0.9}
\setlength{\aboverulesep}{0.2ex}
\setlength{\belowrulesep}{0.3ex}
 
\newcommand{\runtimecell}[2]{%
  \mbox{#1\,{\footnotesize\textcolor{black!65}{(#2)}}}%
}
 
\begin{tabularx}{\linewidth}{@{}l*{3}{>{\centering\arraybackslash}X}@{}}
\toprule
Method & V$^*$Bench & HRBench-4K & HRBench-8K\\
\midrule
Base
& \runtimecell{0.67}{0.25}
& \runtimecell{3.88}{1.21}
& \runtimecell{5.41}{0.22}\\
 
BVS~\cite{bvs}
& \runtimecell{18.74}{5.47}
& \runtimecell{114.11}{28.89}
& \runtimecell{94.83}{29.98}\\
 
HiDe~\cite{hide}
& \runtimecell{3.74}{0.81}
& \runtimecell{12.77}{2.21}
& \runtimecell{16.82}{3.28}\\
 
\rowcolor{black!10}
\textbf{R$^{3}$Sieve}
& \runtimecell{8.20}{1.11}
& \runtimecell{10.33}{0.94}
& \runtimecell{12.07}{0.42}\\
\bottomrule
\end{tabularx}
 
\endgroup
\end{table}

\begin{table}[t]
\centering
\caption{Other VQA benchmarks accuracy. Best results are bold.}
\label{tab:exp_transfer}
\small
\setlength{\tabcolsep}{1.6pt}
\renewcommand{\arraystretch}{0.9}
\setlength{\aboverulesep}{0.2ex}
\setlength{\belowrulesep}{0.3ex}
\begin{tabular*}{\linewidth}{@{\extracolsep{\fill}}l*{8}{c}@{}}
\toprule
& \multicolumn{2}{c}{MMVP}
& \multicolumn{3}{c}{MMStar}
& \multicolumn{3}{c}{MME-RW-Lite}\\
\cmidrule(lr){2-3}
\cmidrule(lr){4-6}
\cmidrule(l){7-9}
Method
& Single & Pair
& Coarse & Fine & All
& Per. & Rea. & All\\
\midrule
Base
& 79.00 & 59.33
& 72.00 & 58.40 & 61.60
& 52.52 & 41.73 & 48.31\\
BVS~\cite{bvs}
& 77.67 & 57.33
& \textbf{74.00} & 55.60 & 62.53
& 57.66 & 44.53 & 52.53\\
HiDe~\cite{hide}
& \textbf{79.33} & 60.00
& 70.00 & 56.40 & 60.27
& 55.60 & 40.80 & 49.82\\
\rowcolor{black!10}
\textbf{R$^{3}$Sieve}
& \textbf{79.33} & \textbf{60.67}
& 70.00 & \textbf{66.00} & \textbf{63.40}
& \textbf{60.14} & \textbf{45.07} & \textbf{54.25}\\
\bottomrule
\end{tabular*}
\end{table}
 
\paragraph{Efficiency.}
\label{sec:exp_efficiency}
We time V$^*$Bench and HRBench-4K/8K , reporting mean runtime per question with standard deviation in parentheses (Table~\ref{tab:efficiency}). Extra passes share a prefix, with fixed preview and crop token budgets, so the runtime of R$^{3}$Sieve grows 1.5 times from V$^*$Bench to HRBench-8K, against 8.1 times for Base (Qwen3-VL-4B) and 4.5 times for HiDe. Its overhead over Base thus falls from 12.2 to 2.2 times, it is faster than HiDe on HRBench-4K and HRBench-8K, and it needs 56.2\% to 90.9\% less time than BVS.

\paragraph{Other VQA tasks.}
\label{sec:exp_general}
We evaluate generalization on MMVP~\cite{mmvp}, MMStar~\cite{mmstar}, and MME-RW-Lite~\cite{mme_realworld} using Qwen3-VL-4B (Table~\ref{tab:exp_transfer}). R$^{3}$Sieve achieves the best or tied-best results in all metrics except coarse MMStar perception, where accuracy drops from 72.00\% to 70.00\%. The largest gains over Base are 7.60 and 7.62 points on fine-grained MMStar and MME-RW-Lite perception, respectively. These tasks rely heavily on local visual details, where suppressing generic responses and preserving complementary evidence are particularly beneficial. Overall, R$^{3}$Sieve generalizes well without substantial degradation on general VQA tasks.

\begin{figure}[t]
    \centering
    \includegraphics[width=\linewidth]{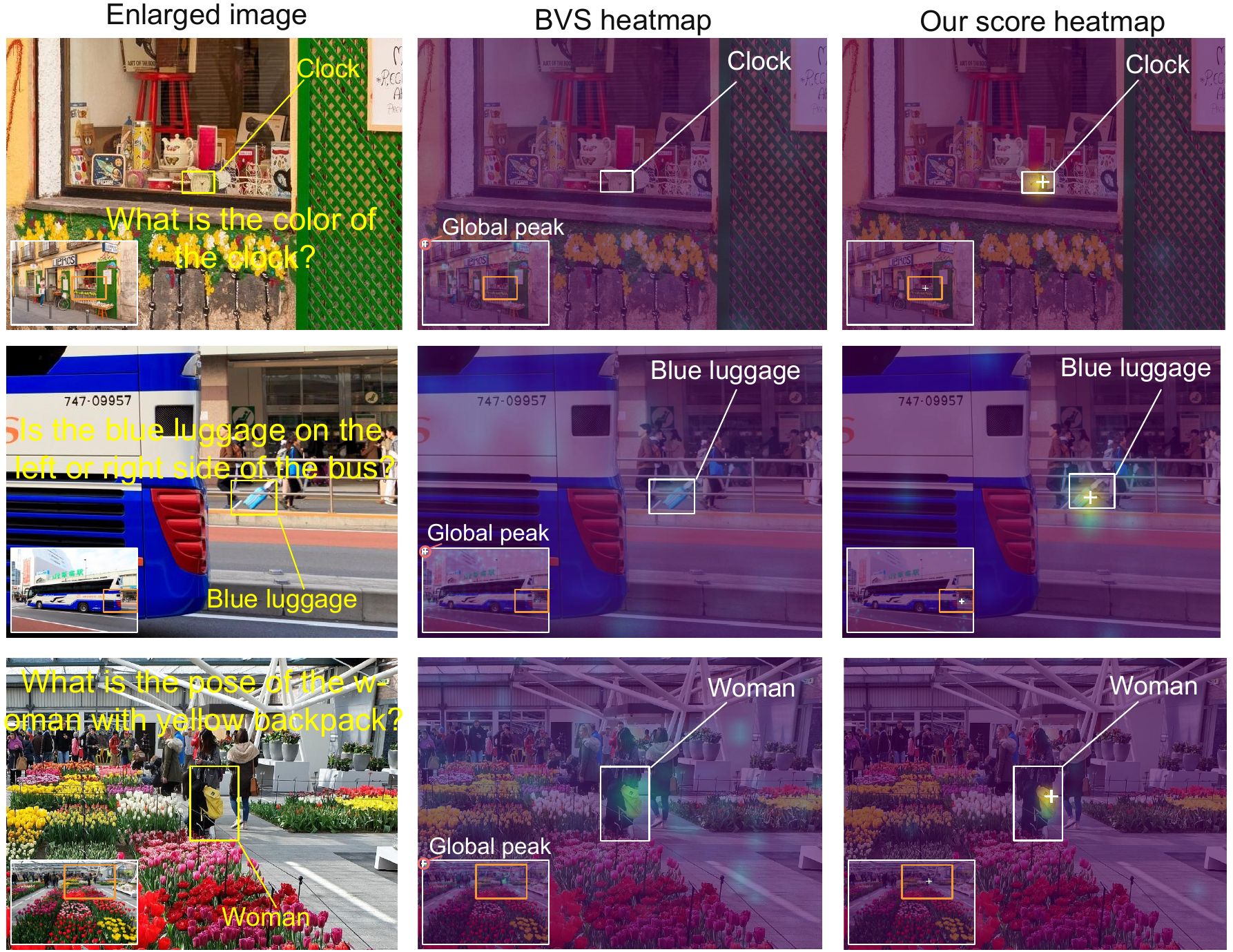}
    \caption{Qualitative comparison of BVS and our contribution scoring. BVS often highlights irrelevant regions, while our method better focuses on the queried targets.}
    \label{fig:score_vis}
\end{figure}

\paragraph{Visualization.}
Fig.~\ref{fig:score_vis} compares BVS with our contribution scoring on three representative examples. While the global peaks of BVS fall on irrelevant regions, our scores better highlight the queried clock, blue luggage, and woman. This suggests that measuring unique regional contributions, rather than response strength alone, helps suppress distracting responses and improves evidence localization and selection.
Complementary crop examples and attention visualizations appear in Supp.~\ref{app:gcs_examples}--\ref{app:attention_visualizations}.

\section{Conclusion}
\label{sec:conclusion}
In high-resolution VQA, models can attend to the right objects yet misjudge their relations as unconditional information obscures evidence. We present R$^{3}$Sieve, a training-free framework that filters unconditional information across target localization, region cropping, and answer generation using the model's own counterfactual responses. At each stage, it retains what the question adds to attention, what a region adds to the selected crops, or what the image adds to an answer. R$^{3}$Sieve achieves the highest overall accuracy among the compared methods across three benchmarks and four backbones. We hope this work offers a new perspective on using visual evidence and advances more generalizable and reliable multimodal reasoning in real-world settings.
\section*{LLM Usage Statement}
During manuscript preparation, the authors used OpenAI Codex
to assist with text polishing, grammar checking, and subsequent
code cleanup. This assistance was based on the manuscript content
and existing experimental results provided by the authors.
AI tools were not used to develop the core algorithm implementation
or to conduct experiments. The authors take full responsibility
for the accuracy and completeness of the final submission.
{
    \small
    \bibliographystyle{ieeenat_fullname}
    \bibliography{main}
}
\clearpage
\setcounter{page}{1}
\maketitlesupplementary
\appendix
\setcounter{figure}{0}
\renewcommand{\thefigure}{S\arabic{figure}}
\renewcommand{\theHfigure}{supp.\arabic{figure}}
\setcounter{table}{0}
\renewcommand{\thetable}{S\arabic{table}}
\renewcommand{\theHtable}{supp.\arabic{table}}

\setcounter{algorithm}{0}
\renewcommand{\thealgorithm}{S\arabic{algorithm}}
\providecommand{\theHalgorithm}{}
\renewcommand{\theHalgorithm}{supp.\arabic{algorithm}}

\section{Overall Algorithm}
\label{app:algorithm}

Algorithm~\ref{alg:r3sieve} summarizes the inference procedure of
R$^{3}$Sieve with a frozen MLLM. SSC extracts the visual residual,
UCS selects complementary windows, and CAV produces the calibrated
answer. Localization and window selection operate on a
preview. The selected windows are then mapped to source-image
coordinates and cropped from the original image. We use a crop budget
of $k=3$ and the fixed settings in Table~\ref{tab:hyperparameters}.

\paragraph{Preview and selection.}
We use the generic prompt ``Describe the image.'' as $q_0$.
Attention is averaged over heads at decoder layer $d$, retaining
the query rows of the corresponding prompt tokens.
$H$ and $N$ share the visual attention mass $m_t$ of each question
token, and $H_0$ uses the corresponding mass under $q_0$.
We set $\epsilon=10^{-12}$.
The operator $\operatorname{OrthoRows}$ constructs the retained
orthonormal row basis as in
Appendix~\ref{app:construction_subspace}.
The candidate set $\Omega$ combines intervals with lengths
$\{1,1/2,1/4,1/8\}$ of each preview-grid axis and a half-length
stride, excluding the full image. Here, $V$ is the number of preview
visual tokens and $|u|$ is the number within window $u$.
The running state $g(j)$ stores the largest detail gain already
provided at location $j$, so each selection counts only the
additional gain.

\paragraph{Answer readout.}
The visual input contains the full image followed by the crops in
selection order. The view captions $m$ and the three branch inputs
follow Tables~\ref{tab:metadata_template} and~\ref{tab:branch_inputs}.
All branches share the same answering instruction.
$\Pi$ contains all $|C|!$ option permutations.
The operator $\operatorname{Logits}_{\mathcal{M}}$ returns the
first-answer-token logits for the valid option letters.
For each order $\pi$, $\operatorname{Align}_{\pi}$ maps the
temperature-normalized log-probabilities from displayed letters
back to answer contents before pooling.
In the pseudocode, $T_q=T_c=T_t$ and $a_q=a_c=a_t$.
At $a_b=0$, pooling uses the mean of the aligned log-probabilities.
All orders within a branch share the common-prefix KV cache,
so the visual input is encoded once.

\begin{algorithm*}[t]
\caption{R$^{3}$Sieve inference.}
\label{alg:r3sieve}
\small
\algrenewcommand{\algorithmicrequire}{\textbf{Input}}
\algrenewcommand{\algorithmicensure}{\textbf{Output}}
\algrenewcommand{\alglinenumber}[1]{\scriptsize #1}
\begin{algorithmic}[1]
\Require Image $I$, question $q$, answer contents $C$, frozen MLLM $\mathcal{M}$, and fixed hyperparameters
\Ensure Predicted answer content $c^\star$

\Statex \textbf{SSC}
\State Extract $A^{\mathrm v},A^{\mathrm n},A_0^{\mathrm v}$ at layer $d$ from preview passes with $q$ and $q_0$
\State $m_t\gets\sum_j A^{\mathrm v}_{tj}$, $m_{0,t}\gets\sum_j A^{\mathrm v}_{0,tj}$
\State $H_{tj}\gets\sqrt{A^{\mathrm v}_{tj}/\max(m_t,\epsilon)}$, $N_{ti}\gets\sqrt{A^{\mathrm n}_{ti}/\max(m_t,\epsilon)}$, $H_{0,tj}\gets\sqrt{A^{\mathrm v}_{0,tj}/\max(m_{0,t},\epsilon)}$
\State $B_0\gets\operatorname{OrthoRows}(H_0)$, $R\gets H-(HB_0^\top)B_0$

\Statex \textbf{UCS}
\State $G_\lambda\gets RR^\top+\lambda NN^\top$, $L\gets\operatorname{Cholesky}(G_\lambda)$
\State Solve $LZ=R$ and set $\phi_\lambda(j)\gets\lVert Z_{\cdot,j}\rVert_2^2$ for all $j$
\State Construct the multiscale candidate set $\Omega$ on the preview grid
\State $\eta(u)\gets(V/|u|)^\rho-1$ for all $u\in\Omega$
\State $S\gets\varnothing$, $g(j)\gets0$ for all $j$
\While{$|S|<k$ and $\Omega\setminus S\neq\varnothing$}
    \State $\Delta(u\mid S)\gets\sum_{j\in u}\phi_\lambda(j)\max\{\eta(u)-g(j),0\}$ for all $u\in\Omega\setminus S$
    \State $u^\star\gets\operatorname*{arg\,max}_{u\in\Omega\setminus S}\Delta(u\mid S)$
    \If{$\Delta(u^\star\mid S)\leq0$}
        \State \textbf{break}
    \EndIf
    \State $S\gets S\cup\{u^\star\}$
    \State $g(j)\gets\max\{g(j),\eta(u^\star)\}$ for all $j\in u^\star$
\EndWhile

\Statex \textbf{CAV}
\State Form $\mathcal{V}$ from $I$ followed by source-image crops of $S$ in selection order
\State Construct the view captions $m$ with source-image crop coordinates
\State $\Pi\gets$ all permutations of $C$, $K\gets|\Pi|$
\For{each $\pi\in\Pi$}
    \State $z_{\mathrm v,\pi}\gets\operatorname{Logits}_{\mathcal{M}}(\mathcal{V},m,q,\pi(C))$
    \State $z_{q,\pi}\gets\operatorname{Logits}_{\mathcal{M}}(q,\pi(C))$
    \State $z_{c,\pi}\gets\operatorname{Logits}_{\mathcal{M}}(m,q,\pi(C))$
    \State $\ell_{b,\pi}\gets\operatorname{Align}_{\pi}\bigl(\operatorname{logsoftmax}(z_{b,\pi}/T_b)\bigr)$ for $b\in\{\mathrm v,q,c\}$
\EndFor
\State $M_b^{(a_b)}(c)\gets\dfrac{1}{a_b}\log\bigl[\dfrac{1}{K}\sum_{\pi\in\Pi}\exp\bigl(a_b\ell_{b,\pi}(c)\bigr)\bigr]$ for $b\in\{\mathrm v,q,c\}$ and $c\in C$
\State $s(c)\gets M_{\mathrm v}^{(a_v)}(c)-\beta M_q^{(a_t)}(c)+\gamma\bigl[M_c^{(a_t)}(c)-M_q^{(a_t)}(c)\bigr]$ for all $c\in C$
\State \Return $c^\star\gets\operatorname*{arg\,max}_{c\in C}s(c)$
\end{algorithmic}
\end{algorithm*}

\section{Attention Coverage in Incorrect Answers}
\label{app:attention_errors}

\subsection{Coverage Definitions and Results}
\label{app:attention_coverage_definitions}

\paragraph{Evaluation setup.}
We evaluate Qwen3-VL-4B/8B and InternVL3.5-4B/8B on all
191 questions of V$^*$Bench, predicting the valid option letter
with the highest next-token logit.
Target boxes come from the original per-image JSON annotations
of V$^*$Bench~\cite{vstar}.
For the forward pass predicting the first answer token, we
collect attention from the final prompt position, average over
heads, and map visual responses from all 36 decoder layers to
source-image coordinates.
At each layer, we retain the top 10\% of visual grid cells.
A target is covered when these cells cover at least 50\% of
its annotated box area, its mean attention density is at least
twice the image average, and the image receives at least 1\%
of total attention.
Boundary cells contribute according to their overlapping area
with the target box.

\paragraph{Coverage definitions.}
Let $h_{ik}^{(\ell)}\in\{0,1\}$ indicate whether annotated
target $k$ of question $i$ satisfies these conditions at layer
$\ell$, with $K_i$ annotated targets and $L=36$ layers.
We compare cross-layer and same-layer coverage
\begin{equation}
\label{eq:app_attention_coverage_events}
\begin{aligned}
C_i^{\mathrm{any}}
&=\min_{1\leq k\leq K_i}
  \max_{1\leq\ell\leq L}h_{ik}^{(\ell)},\\
C_i^{\mathrm{same}}
&=\max_{1\leq\ell\leq L}
  \min_{1\leq k\leq K_i}h_{ik}^{(\ell)}.
\end{aligned}
\end{equation}
$C_i^{\mathrm{any}}$ allows different targets to satisfy the
criterion at different layers.
$C_i^{\mathrm{same}}$ requires all annotated targets to satisfy
it simultaneously within a common layer.
These layers are depths within the same forward pass.
The two definitions coincide for a single annotated target.

Table~\ref{tab:app_attention_coverage_definitions} reports
coverage among each model's own incorrect answers.
Each entry gives the covered-error count divided by the total
error count, followed by the percentage.
Alongside all 191 questions, we separately analyze the 54 spatial
questions with both queried targets annotated.
The remaining 22 spatial questions have one annotated target
each, and their coverage statistics concern the available box.

\begin{table}[t]
\centering
\small
\caption{Geometric attention coverage among incorrect answers.}
\setlength{\tabcolsep}{6pt}
\renewcommand{\arraystretch}{1.10}
\begin{tabular}{lrrr}
\toprule
Model & Errors & $C^{\mathrm{any}}$ & $C^{\mathrm{same}}$\\
\midrule
\multicolumn{4}{l}{\textit{All 191 questions}}\\
Qwen3-VL-4B    & 35 & 28/35 (80.0\%) & 27/35 (77.1\%)\\
Qwen3-VL-8B    & 33 & 31/33 (93.9\%) & 30/33 (90.9\%)\\
InternVL3.5-4B & 52 & 44/52 (84.6\%) & 43/52 (82.7\%)\\
InternVL3.5-8B & 51 & 41/51 (80.4\%) & 40/51 (78.4\%)\\
\midrule
\multicolumn{4}{l}{\shortstack[l]{
    \textit{54 spatial questions with}\\
    \textit{both queried targets annotated}
}}\\
Qwen3-VL-4B    & 15 & 10/15 (66.7\%) & 9/15 (60.0\%)\\
Qwen3-VL-8B    & 13 & 11/13 (84.6\%) & 10/13 (76.9\%)\\
InternVL3.5-4B & 13 & 9/13 (69.2\%)  & 8/13 (61.5\%)\\
InternVL3.5-8B & 15 & 11/15 (73.3\%) & 10/15 (66.7\%)\\
\bottomrule
\end{tabular}
\label{tab:app_attention_coverage_definitions}
\end{table}

\begin{figure*}[t]
\centering
\includegraphics[width=\linewidth]{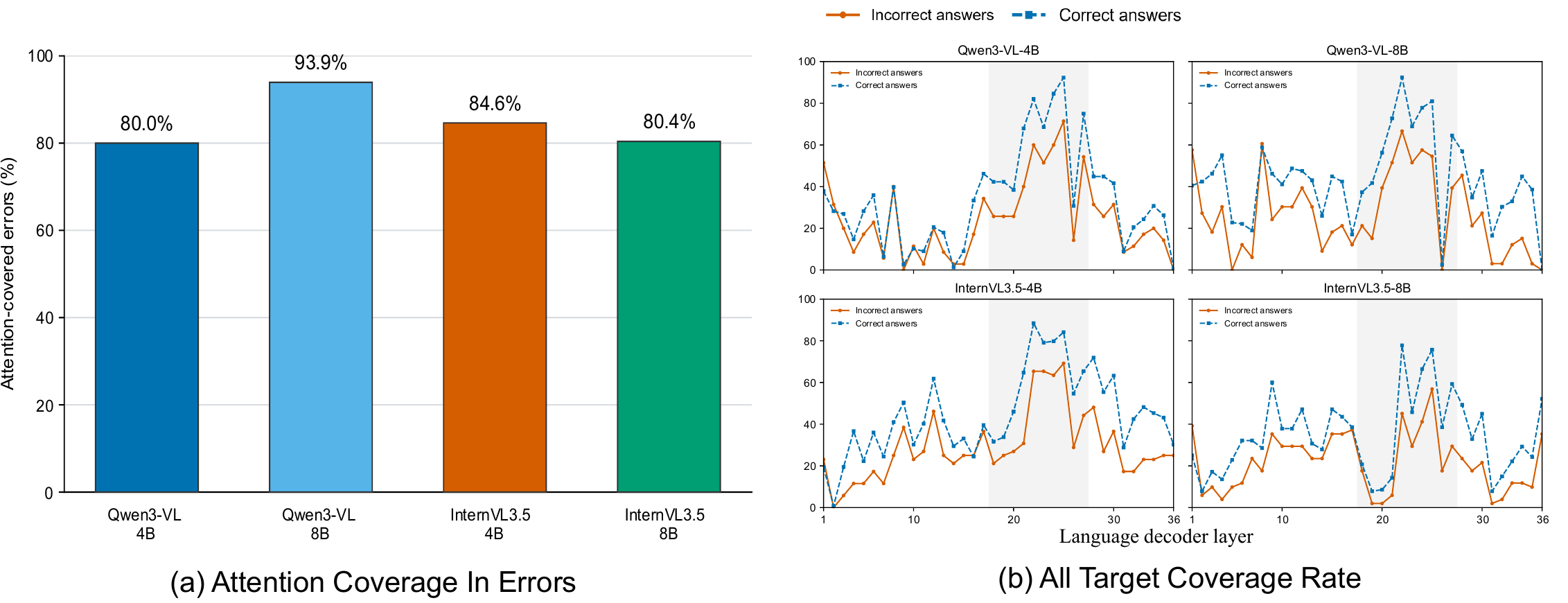}
\caption{Attention coverage on V$^*$Bench.
(a) Cross-layer coverage among incorrect answers.
(b) Simultaneous coverage of all annotated targets at each
decoder layer for correct and incorrect answers.
Shading marks layers 18--27.}
\label{fig:app_attention_errors}
\end{figure*}

\paragraph{Results and analysis.}
Even under the stricter coverage requirement, most incorrect
answers remain covered
(Table~\ref{tab:app_attention_coverage_definitions}).
Requiring a common layer removes only one covered error per
model relative to cross-layer coverage, so the high coverage
rate persists when targets must be covered simultaneously.
This pattern also holds on spatial questions with both queried
targets annotated, showing that the gap between coverage and
correctness persists under simultaneous multi-target coverage.
Figure~\ref{fig:app_attention_errors}(a) reports cross-layer
coverage, while Fig.~\ref{fig:app_attention_errors}(b) shows
simultaneous coverage at each depth for correct and incorrect
answers.
Geometric coverage measures spatial alignment between attention
and target regions, while correct answers additionally require
identifying and using the relevant visual content.
These results motivate controlled interventions on visual
evidence acquisition and answer readout.

\subsection{Controlled Target Crops and Answer Readout}
\label{app:oracle_readout_diagnostic}

\paragraph{Controlled comparison.}
We evaluate Qwen3-VL-4B on all 191 V$^*$Bench questions,
comparing automatic GCS crops with crops derived from annotated
target boxes. Each annotated box is expanded by a factor of two
to retain local context. The two conditions retain the same
global image and match the number of detail views and actual
visual-token counts for each question. The global view uses up
to 4,096 tokens and each detail view uses 1,024 tokens.
These diagnostic budgets are separate from the full-method
configuration. Within each crop condition, we hold the images
fixed and compare the original option order with complete
permutation pooling aligned by answer content.
Pooling uses $a_{\mathrm v}=0.375$ and unit temperature.

\begin{table}[t]
\centering
\small
\caption{Accuracy under controlled crop inputs (\%).}
\setlength{\tabcolsep}{6pt}
\begin{tabular}{lrr}
\toprule
Crop input & Original order & Pooled\\
\midrule
Automatic        & 84.82 & 86.91\\
Annotated target & 88.48 & 91.62\\
\bottomrule
\end{tabular}

\label{tab:app_oracle_readout}
\end{table}

\paragraph{Results and analysis.}
Option presentation affects predictions even when target crops
are explicitly supplied. With annotated-target images fixed,
16 questions change their predicted answer content under option
permutations. Pooling corrects seven original-order errors and
introduces one new error
(Table~\ref{tab:app_oracle_readout}).
This effect also appears on the 54 spatial questions with both
queried targets annotated. Eleven are sensitive to option
permutations, and pooling increases the correct count from
38 to 43 with no regressions.
Here, both target views and the global spatial context remain
available throughout the comparison. The remaining sensitivity
therefore concerns how the same visual input is processed under
different option presentations. Better target crops and
order-robust readout provide complementary improvements in this
diagnostic.

\subsection{Target Deletion and Background Interference}
\label{app:target_background_interventions}

\paragraph{Dependence on target regions.}
We evaluate all 115 attribute questions with Qwen3-VL-4B using
a single global image and the original option order.
We replace the target region with gray and compare it with
deletion of an equally sized region outside the target.
Mask area and visual-token count are matched for each question.
We measure the ground-truth option's logit margin over the
strongest alternative. Target deletion reduces this margin
by an additional 13.32 relative to control deletion,
with a 95\% paired bootstrap interval of $[11.53,15.08]$.
The additional reduction is 1.47 on the 15 originally incorrect
questions, with an interval of $[0.69,2.38]$.
Thus, answer scores depend on target-region content even for
some incorrect predictions. Since target deletion removes
answer evidence, its effect is assessed through score changes.

\paragraph{Effect of background content.}
We preserve the original pixels within target boxes expanded
by a factor of 1.25 and blur the remaining background.
Background suppression corrects more answers than it disrupts
(Table~\ref{tab:app_background_intervention}), while the target
pixels remain unchanged. This shows that background content can
affect decisions even when the local answer evidence is
preserved. Suppression can also make targets easier to localize.
We therefore add a follow-up control that keeps annotated-target
detail views fixed and changes only the global background.
This follow-up includes all 115 attribute questions and was
specified after observing the single-image result.
With explicit target details available, background processing improves accuracy by 0.87 percentage points. The mean ground-truth margin increases by 1.40, with a 95\% confidence interval of $[1.06,1.75]$.
The smaller accuracy change occurs from an already high baseline,
while the score change shows that background sensitivity
persists after target details are supplied.

\begin{table}[t]
\centering
\small
\caption{Correct answers under background processing (\%).}
\setlength{\tabcolsep}{6pt}
\begin{tabular}{lrr}
\toprule
Target detail views & Original & Blurred\\
\midrule
Absent & 86.96 & 93.91\\
Fixed  & 97.39 & 98.26\\
\bottomrule
\end{tabular}

\label{tab:app_background_intervention}
\end{table}

\paragraph{Paired analysis.}
Intervals use 5,000 paired bootstrap resamples of questions.
The accuracy differences are $+6.96$ percentage points
with interval $[0.87,13.04]$ for the single-image comparison
and $+0.87$ with interval $[-1.74,4.35]$ when target details
are fixed. Two-sided exact McNemar tests give $p=0.0703$
for annotated-crop pooling, $p=0.0574$ for single-image
background processing, and $p=1.0000$ with fixed target details.

Together, these interventions show that model scores and
decisions respond to target evidence, background content, and
option presentation.

\section{Experimental Details of the Analysis}
\label{app:comparison-details}

\subsection{Variants for Target Localization}
\label{app:background-processing}
The four variants in Sec.~\ref{sec:visual_interference} share the visual attention of all question tokens, and the three variants with a reference also share the attention that the prompt \textit{``Describe the image.''} elicits on the same image. All variants use the same spatial normalization, region selection, and answer-generation pipeline, so they differ only in how they remove the unconditional component.

\paragraph{Raw Attention.}
We average the visual attention over the question tokens and use the resulting map $\bar a$ as the region score without a reference.

\paragraph{Pointwise Ratio.}
Following the relative attention of ViCrop~\cite{vicrop}, we average the question and generic-prompt attention over their text tokens and divide $\bar a$ by $\bar a_0$ elementwise. The ratio removes $\log\bar a_0$ in the log domain and favors locations whose response rises under the question.

\paragraph{Pointwise Difference.}
We adopt the smoothing and subtraction steps of token-wise attention decoupling in HiDe~\cite{hide}, with the generic-prompt attention as the reference. Each question-token and generic-prompt map is smoothed with a Gaussian kernel of standard deviation $\sigma=3$ preview-grid cells, using reflect boundary handling and truncation at $4\sigma$ (a separable $25\times25$ spatial kernel), and then min--max normalized. The processed generic-prompt maps are averaged into $\bar a_0$. We subtract $\bar a_0$ from each processed question-token map, clip negative values to zero, normalize again, and sum the maps into the final score.

\paragraph{Subspace Projection.}
We take element-wise square roots of the question and generic-prompt visual attention to obtain $H$ and $H_0$, build the row-orthonormal basis $B_0$ of Appendix~\ref{app:construction_subspace}, and score each location by $d_j=\lVert R_{:,j}\rVert_2^2$ with $R$ from Eq.~\eqref{eq:residual}. The basis keeps every spatial pattern of the generic responses, so the projection removes all of them, whereas the pointwise variants compare against a single averaged map.

\subsection{Metrics for Visual Evidence Selection}
\label{app:selection_metrics}

We evaluate local selections using minimum target coverage
and enrichment, and report answer accuracy with the full
image and selected crops.
For question $i$, let $I_i$ denote the original image region
and $G_{ik}$ the annotated target boxes, where
$k=1,\ldots,K_i$.
Let $G_i=\bigcup_{k=1}^{K_i}G_{ik}$ be their union and
$U_{im}(f)$ the union of local regions selected by method $m$
at nominal area budget $f$.
The notation $|\cdot|$ denotes area.
Target annotations are used only for evaluation, and the
geometric metrics measure the coverage of the local crops.

\paragraph{Crop-area budget.}
The actual selected area fraction is
\begin{equation}
\label{eq:metric_actual_area}
f_{im}^{\mathrm{actual}}
=\frac{|U_{im}(f)|}{|I_i|}.
\end{equation}
This diagnostic selects one window for attribute questions
and two non-overlapping windows for spatial-relation questions.
Given an $h_i\times w_i$ preview grid and $L_i$ windows,
each window has grid dimensions
\begin{equation}
\label{eq:metric_window_dimensions}
\begin{aligned}
d_{h,i}
&=\max\left(1,\operatorname{round}
\left(h_i\sqrt{f/L_i}\right)\right),\\
d_{w,i}
&=\max\left(1,\operatorname{round}
\left(w_i\sqrt{f/L_i}\right)\right).
\end{aligned}
\end{equation}
For each method, we enumerate all grid-aligned windows of
this size within the preview grid and score each window
by the sum of its cell scores.
We greedily select the highest-scoring window, repeating
until $L_i$ windows have been selected while requiring each
new window to be disjoint from all previously selected windows.
Ties are resolved in row-major order of the top-left coordinates.
Each budget is evaluated independently, so selected regions
need not be nested.

The actual selected-cell fraction is
$L_i d_{h,i}d_{w,i}/(h_iw_i)$, which differs slightly from the
nominal budget $f$ on the horizontal axes because of grid rounding.
For each question and budget, all methods select the same
number of grid cells and use the same actual number of
crop-encoding tokens.

\paragraph{Minimum target coverage.}
We compute the covered fraction of each annotated target
and take the minimum within each question,
\begin{equation}
\label{eq:metric_minimum_coverage}
\kappa_{im}(f)
=\min_{1\leq k\leq K_i}
\frac{|G_{ik}\cap U_{im}(f)|}{|G_{ik}|}.
\end{equation}
This measures the retained area fraction of the
least-covered annotated target.
For a question with one annotated target, it reduces to
that target's coverage.
We average the per-question minima,
\begin{equation}
\label{eq:metric_mean_coverage}
\overline{\kappa}_m(f)
=\frac{1}{N}\sum_{i=1}^{N}\kappa_{im}(f).
\end{equation}
The figure reports this mean as a percentage, with
$N=191$ questions in this experiment.

\paragraph{Area computation.}
Geometric metrics are computed on the equal-area preview
grid mapped to the original image.
For grid cell $Q_{ij}$, we compute its fractional overlap
with each target box and their union,
\begin{equation}
\label{eq:metric_fractional_overlap}
a_{ikj}=\frac{|Q_{ij}\cap G_{ik}|}{|Q_{ij}|},
\qquad
b_{ij}=\frac{|Q_{ij}\cap G_i|}{|Q_{ij}|}.
\end{equation}
Let $z_{imj}(f)\in\{0,1\}$ indicate whether cell $j$ is
selected.
The metrics are then computed as
\begin{equation}
\label{eq:metric_grid_computation}
\begin{aligned}
\kappa_{im}(f)
&=\min_k
\frac{\sum_j a_{ikj}z_{imj}(f)}
{\sum_j a_{ikj}},\\
e_{im}(f)
&=\frac{
\sum_j b_{ij}z_{imj}(f)/\sum_j z_{imj}(f)
}{
\sum_j b_{ij}/V_i
},
\end{aligned}
\end{equation}
where $V_i=h_iw_i$.
This computation preserves partial coverage when target
boxes cross grid boundaries.
Overlapping target boxes and selected regions are handled
through their respective unions.

\paragraph{Answer accuracy.}
Let $\hat y_{im}(f)$ and $y_i$ denote the predicted and
ground-truth answers.
Accuracy is
\begin{equation}
\label{eq:metric_answer_accuracy}
\operatorname{Acc}_m(f)
=\frac{1}{N}\sum_{i=1}^{N}
\mathbf 1[\hat y_{im}(f)=y_i].
\end{equation}
Here, $\mathbf{1}[\cdot]$ denotes the indicator function, which equals one when its argument is true and zero otherwise. All methods answer using the full image and selected crops, and accuracy is reported as a percentage.

\section{Additional Method Details}
\label{app:method_details}

\subsection{Construction of the Shared Subspace}
\label{app:construction_subspace}
We eigendecompose the generic-prompt Gram matrix as $H_0H_0^\top=U\Lambda U^\top$, keep the eigenvalues larger than $10^{-5}$ times the largest one, and denote the kept eigenvectors and eigenvalues by $U_r$ and $\Lambda_r$. The orthonormal row basis is
\begin{equation}
\label{eq:freeeyes_basis}
B_0=\Lambda_r^{-1/2}U_r^\top H_0,
\end{equation}
which satisfies $B_0B_0^\top=I$, so the residual in Eq.~\eqref{eq:residual} is orthogonal to the retained generic-prompt subspace.

\subsection{Token Contributions as Leverage Scores}
\label{app:construction_token}
Let $X_\lambda=[R,\sqrt{\lambda}N]$, so that $G_\lambda=X_\lambda X_\lambda^\top$. When $G_\lambda$ is invertible,
\begin{equation}
\label{eq:freeeyes_joint_projection}
P_\lambda=X_\lambda^\top(X_\lambda X_\lambda^\top)^{-1}X_\lambda
\end{equation}
is the orthogonal projector onto the row space of $X_\lambda$, and the visual token scores are its first $V$ diagonal entries, $\phi_\lambda(j)=(P_\lambda)_{jj}$, which are the statistical leverage scores of the visual columns~\cite{mahoney}. Each score thus measures how much of a visual column the other visual and nonvisual columns cannot reproduce. The nonvisual keys include system text, instructions, and question tokens. Their hidden states can carry visual information, so they join the other visual columns as the counterfactual reference of Eq.~\eqref{eq:contribution}.

\begin{table}[t]
\centering
\begingroup
\small
\caption{Fixed view captions and shared answering instruction. Crop coordinates refer to the source image.}
\label{tab:metadata_template}
\setlength{\tabcolsep}{0pt}
\renewcommand{\arraystretch}{1.12}
\begin{tabularx}{\columnwidth}{@{}>{\raggedright\arraybackslash}X@{}}
\toprule
\textbf{Full-image caption}\\[2pt]
{\ttfamily Full source image.}\\
\midrule
\textbf{Crop caption} \normalfont(repeated for each crop)\\[2pt]
{\ttfamily A detail view of the same image, source rectangle
[l, t, r, b] in [left, top, right, bottom] coordinates on a
0-1000 grid. Use the full image to determine global spatial
relationships.}\\
\midrule
\textbf{Shared answering instruction}\\[2pt]
{\ttfamily Answer with only the option letter.}\\
\bottomrule
\end{tabularx}
\endgroup

\end{table}

\begin{table}[t]
\centering
\begingroup
\small
\caption{Inputs to the three branches. All branches also receive the shared answering instruction in Table~\ref{tab:metadata_template}.}
\label{tab:branch_inputs}
\setlength{\tabcolsep}{5pt}
\renewcommand{\arraystretch}{1.12}
\begin{tabularx}{\columnwidth}{@{}>{\raggedright\arraybackslash}Xccc@{}}
\toprule
\textbf{Branch} & \textbf{Images} & \textbf{Captions} &
\shortstack{\textbf{Question}\\\textbf{+ options}}\\
\midrule
Visual        & Yes & Yes & Yes\\
Metadata-only & No  & Yes & Yes\\
Question-only & No  & No  & Yes\\
\bottomrule
\end{tabularx}
\endgroup
\end{table}

\subsection{Metadata template.}
\label{metadata_template}
Table~\ref{tab:branch_inputs} and Table~\ref{tab:metadata_template} lists the fixed view captions,
the shared answering instruction, and the inputs to each branch.
For source-image dimensions $W\times H$ and source-pixel crop
$(x_0,y_0,x_1,y_1)$, the caption encodes the source rectangle as
$[l,t,r,b]$ on a 0--1000 grid, where
\[
\begin{aligned}
l&=\operatorname{rnd}(1000x_0/W), &
t&=\operatorname{rnd}(1000y_0/H),\\
r&=\operatorname{rnd}(1000x_1/W), &
b&=\operatorname{rnd}(1000y_1/H).
\end{aligned}
\]
Here, $\operatorname{rnd}$ rounds to the nearest integer, with ties
to even.
The visual branch presents the full image followed by the crops in
selection order, with each caption immediately before its image.
The metadata-only branch retains the captions in the same order
but omits all images; the question-only branch omits both images
and captions.
All branches receive the question and options followed by the
shared answering instruction in Table~\ref{tab:metadata_template}.

\section{Proofs of Propositions}
\label{app:proofs}
 
\subsection{Orthogonal Residual}
\label{app:proof_subspace}
\begin{proof}[Proof of Proposition~\ref{prop:subspace}]
Let $P=B_0^\top B_0$. Since $B_0B_0^\top=I$, $P$ is symmetric with $P^2=P$, so it is the orthogonal projector onto the row space of $B_0$, and $R=H(I-P)$.
 
\emph{Orthogonality.} $RB_0^\top=HB_0^\top-HB_0^\top B_0B_0^\top=0$.
 
\emph{Best approximation.} The set $\mathcal{X}=\{X: XB_0^\top=0\}=\{X: XP=0\}$ is a linear subspace that contains $R$ by orthogonality. For any $X\in\mathcal{X}$, we have $H-X=HP+(R-X)$ and
\begin{equation*}
\begin{aligned}
\langle HP,R-X\rangle_F&=\operatorname{tr}\bigl(PH^\top(R-X)\bigr)\\
&=\operatorname{tr}\bigl(H^\top(R-X)P\bigr)=0,
\end{aligned}
\end{equation*}
because $(R-X)P=0$. Hence $\lVert H-X\rVert_F^2=\lVert HP\rVert_F^2+\lVert R-X\rVert_F^2\geq\lVert H-R\rVert_F^2$, with equality only for $X=R$.
 
\emph{Energy split.} In the same way, $\langle R,HP\rangle_F=\operatorname{tr}(HPR^\top)=\operatorname{tr}\bigl(H(RP)^\top\bigr)=0$, so $\lVert H\rVert_F^2=\lVert R\rVert_F^2+\lVert HP\rVert_F^2$. Moreover,
\begin{equation*}
\begin{aligned}
\lVert HP\rVert_F^2&=\operatorname{tr}\bigl(B_0^\top B_0H^\top HB_0^\top B_0\bigr)\\
&=\operatorname{tr}\bigl(B_0H^\top HB_0^\top\bigr)=\lVert HB_0^\top\rVert_F^2,
\end{aligned}
\end{equation*}
which gives the energy identity.
\end{proof}
 
\subsection{Discount of Repeated Responses}
\label{app:proof_repeat}
\begin{proof}[Proof of Proposition~\ref{prop:repeat}]
The bound is a standard property of leverage scores~\cite{mahoney}, and we prove it in our notation. Let $A_j=\sum_{i=1}^m\alpha_i^2$. Since $RR^\top=\sum_kr_kr_k^\top$ and $NN^\top\succeq0$, we have
\begin{equation*}
G_\lambda\succeq r_jr_j^\top+\sum_{i=1}^mr_{k_i}r_{k_i}^\top=(1+A_j)\,r_jr_j^\top,
\end{equation*}
so $w^\top G_\lambda w\geq(1+A_j)(w^\top r_j)^2$ for every $w$. Taking $w=G_\lambda^{-1}r_j$ gives $\phi_\lambda(j)\geq(1+A_j)\,\phi_\lambda(j)^2$, and hence $\phi_\lambda(j)\leq1/(1+A_j)$ whenever $r_j\neq0$. The bound holds trivially when $r_j=0$.
\end{proof}
 
\paragraph{Identity in Eq.~\eqref{eq:contribution}.}
We also verify that $\phi_\lambda(j)$ grows with the energy that only location $j$ adds, as stated for $\lambda=1$ in Eq.~\eqref{eq:contribution}.
 
\begin{lemma}[Generalized Rayleigh quotient]
\label{lem:rayleigh}
For a positive definite matrix $A$ and a vector $r\neq0$,
\begin{equation*}
\max_{w\neq0}\frac{(w^\top r)^2}{w^\top Aw}=r^\top A^{-1}r,
\end{equation*}
and the maximum is attained at $w=A^{-1}r$.
\end{lemma}
\begin{proof}
Write $v=A^{1/2}w$. By the Cauchy--Schwarz inequality,
\begin{equation*}
\begin{aligned}
(w^\top r)^2&=\bigl(v^\top A^{-1/2}r\bigr)^2\leq\lVert v\rVert_2^2\,\lVert A^{-1/2}r\rVert_2^2\\
&=(w^\top Aw)(r^\top A^{-1}r),
\end{aligned}
\end{equation*}
with equality when $v$ is parallel to $A^{-1/2}r$, that is, when $w$ is a multiple of $A^{-1}r$.
\end{proof}
 
Let $G_{\lambda,-j}=G_\lambda-r_jr_j^\top$ be positive definite and $q_j=\max_{w\neq0}(w^\top r_j)^2/(w^\top G_{\lambda,-j}w)$. By Lemma~\ref{lem:rayleigh}, $q_j=r_j^\top G_{\lambda,-j}^{-1}r_j$. Since $G_\lambda=G_{\lambda,-j}+r_jr_j^\top$, the Sherman--Morrison formula gives
\begin{equation*}
G_\lambda^{-1}=G_{\lambda,-j}^{-1}-\frac{G_{\lambda,-j}^{-1}r_jr_j^\top G_{\lambda,-j}^{-1}}{1+q_j},
\end{equation*}
so that $\phi_\lambda(j)=r_j^\top G_\lambda^{-1}r_j=q_j-q_j^2/(1+q_j)=q_j/(1+q_j)$, which is increasing in $q_j$ and reduces to Eq.~\eqref{eq:contribution} for $\lambda=1$.
 
\subsection{Approximation Guarantee of GCS}
\label{app:gcs-guarantee}
 
\paragraph{Setup.}
Let $\Omega$ be a finite set of candidate windows and let
$k\geq 1$ be the window-count budget.
Assume that the token contribution scores
$w_j:=\phi_\lambda(j)$ are finite and nonnegative, and that
$\Omega$, $w_j$, and the window detail gains $\eta(u)$
remain fixed throughout selection.
We consider
\begin{equation}
\max_{S\subseteq\Omega,\ |S|\leq k} F(S),
\qquad
F(S)=\sum_{j=1}^{V}w_j g_S(j),
\label{eq:gcs-optimization}
\end{equation}
where
\begin{equation}
g_S(j)=
\max\bigl(
\{\eta(u):u\in S,\ j\in u\}\cup\{0\}
\bigr).
\label{eq:gcs-coverage-app}
\end{equation}
The only constraint is the window-count budget.
At each step, GCS selects an unselected window with the
largest exact marginal gain.
It stops when the budget is reached, the candidate set is
exhausted, or no positive marginal gain remains.
 
\paragraph{Statement.}
We restate Proposition~\ref{prop:gcs} in full.
The objective $F$ is normalized, monotone, and submodular.
GCS makes at most $\min\{k,|\Omega|\}$ selections.
Let $S_{\mathrm G}$ be its returned set and let
\begin{equation}
S^\star\in
\arg\max_{S\subseteq\Omega,\ |S|\leq k}F(S).
\end{equation}
Then
\begin{equation}
\begin{aligned}
F(S_{\mathrm G})
&\geq
\left[1-\left(1-\frac{1}{k}\right)^k\right]F(S^\star)\\
&\geq (1-1/e)F(S^\star).
\end{aligned}
\label{eq:gcs-approximation}
\end{equation}
In particular, for $k=3$,
\begin{equation}
F(S_{\mathrm G})\geq \frac{19}{27}F(S^\star).
\label{eq:gcs-three-window-bound}
\end{equation}
 
\begin{proof}[Proof of Proposition~\ref{prop:gcs}]
By definition, $g_{\varnothing}(j)=0$, so
$F(\varnothing)=0$.
For any $A\subseteq B\subseteq\Omega$,
the pointwise maximum satisfies $g_A(j)\leq g_B(j)$.
Since $w_j\geq 0$, we obtain $F(A)\leq F(B)$.
Thus, $F$ is normalized and monotone.
 
For any $u\in\Omega\setminus S$, its marginal gain is
\begin{equation}
\begin{aligned}
\Delta(u\mid S)
&=F(S\cup\{u\})-F(S)\\
&=\sum_{j\in u}w_j[\eta(u)-g_S(j)]_+,
\end{aligned}
\label{eq:gcs-marginal-app}
\end{equation}
where $[x]_+=\max\{x,0\}$.
For $A\subseteq B$ and $u\in\Omega\setminus B$,
$g_A(j)\leq g_B(j)$ implies
\begin{equation}
\begin{aligned}
\Delta(u\mid A)
&=\sum_{j\in u}w_j[\eta(u)-g_A(j)]_+\\
&\geq\sum_{j\in u}w_j[\eta(u)-g_B(j)]_+\\
&=\Delta(u\mid B).
\end{aligned}
\end{equation}
Hence, $F$ satisfies the diminishing-returns condition
and is submodular.
 
Each accepted selection adds a previously unselected
window with strictly positive marginal gain.
The objective therefore strictly increases at every
accepted selection.
Since the number of selections is bounded by both $k$
and $|\Omega|$, GCS terminates after at most
$\min\{k,|\Omega|\}$ selections.
 
First suppose that GCS completes $k$ selections.
Let $S_t$ denote the set after $t$ selections and let
$u_t$ be the next greedily selected window.
By monotonicity, submodularity, and the greedy rule,
\begin{equation}
\begin{aligned}
F(S^\star)-F(S_t)
&\leq F(S_t\cup S^\star)-F(S_t)\\
&\leq
\sum_{u\in S^\star\setminus S_t}\Delta(u\mid S_t)\\
&\leq k\,\Delta(u_t\mid S_t).
\end{aligned}
\label{eq:gcs-gap-bound}
\end{equation}
The second inequality follows by adding the windows
in $S^\star\setminus S_t$ sequentially and applying
diminishing returns.
The last inequality uses $|S^\star|\leq k$ and the
maximality of the greedy marginal gain.
 
Writing $D_t=F(S^\star)-F(S_t)$ gives
\begin{equation}
D_{t+1}\leq
\left(1-\frac{1}{k}\right)D_t.
\end{equation}
Since $S_0=\varnothing$ and $F(S_0)=0$,
\begin{equation}
D_k\leq
\left(1-\frac{1}{k}\right)^kF(S^\star).
\end{equation}
Rearranging yields the first inequality in
Eq.~\eqref{eq:gcs-approximation}.
 
Now suppose that GCS stops before reaching the budget
because no positive marginal gain remains.
Monotonicity implies that all remaining marginal gains
are zero.
Consequently,
\begin{equation}
\begin{aligned}
0
&\leq F(\Omega)-F(S_t)\\
&\leq
\sum_{u\in\Omega\setminus S_t}\Delta(u\mid S_t)
=0.
\end{aligned}
\end{equation}
Thus, $F(S_t)=F(\Omega)$.
Since $S_t$ is feasible, it also attains the
budget-constrained optimum.
The same conclusion holds if the candidate set is
exhausted.
Therefore, early stopping preserves the guarantee.
 
Finally, $(1-1/k)^k\leq e^{-1}$ gives the
$1-1/e$ bound.
For $k=3$, the finite-budget factor is
$1-(2/3)^3=19/27$.
\end{proof}
 
\subsection{Invariance to Option Position}
\label{app:proof_order}
\begin{proof}
Reordering the supplied options leaves the set of all
displayed arrangements unchanged. Mapping scores back
to contents therefore produces the same score multiset
for each content. Symmetric pooling preserves these
scores for every exponent.

For the ranking result, fix $g(c)>g(d)$.
For each order $\pi$, let $\pi^{cd}$ exchange the
positions of $c,d$ while leaving other contents fixed.
Write
\[
r=e^{g(c)},\quad s=e^{g(d)},\quad
u=e^{b(p_\pi(c))},\quad v=e^{b(p_\pi(d))},
\]
and let $A\ge0$ denote the sum of the other contents'
unnormalized weights. Then $r>s$ and
\[
P_\pi(c)=\frac{ru}{ru+sv+A},
\qquad
P_{\pi^{cd}}(d)=\frac{su}{rv+su+A}.
\]
After cancelling the common positive factor $u$,
the cross-multiplied difference is
\begin{align*}
&r(rv+su+A)-s(ru+sv+A)\\
&\qquad=(r-s)\bigl((r+s)v+A\bigr)>0.
\end{align*}
Hence $\ell_\pi(c)>\ell_{\pi^{cd}}(d)$ for every $\pi$.

Complete enumeration makes the map $\pi\mapsto\pi^{cd}$ a bijection on the set of all option permutations. Mellowmax is strictly increasing in every finite input score for every real exponent, including its mean extension at $a=0$. Therefore, $M^{(a)}(c)>M^{(a)}(d)$. When $g(c)=g(d)$, the same pairing gives equality. Exchanging $c$ and $d$ proves the remaining case.
\end{proof}

\section{Evaluation Details}
\label{app:experiment_details}

\subsection{Visual Tokens Budget}
\label{app:visual_tokens}

For both Qwen3-VL backbones, the complete image is processed with an upper limit of $16{,}384$ visual tokens. Each crop is resized toward a $3{,}072$-token target under the same dimension alignment. Both InternVL3.5-4B and InternVL3.5-8B use native dynamic tiling with at most $36$ spatial tiles of $448\times448$ pixels for the complete image, adding one thumbnail when more than one spatial tile is used. Each spatial tile and thumbnail contributes $256$ visual tokens, giving a full-image maximum of $(36+1)\times256=9{,}472$ tokens. The per-crop cap is $3{,}072$ visual tokens, including any thumbnail.

\subsection{Hyperparameter Settings}
\label{app:hyperparameters}

We select the hyperparameters for each backbone using 200 questions.
These comprise 41 questions drawn from the additional OCR and
GPT4V-hard subsets of V$^{*}$Bench~\cite{vstar}, beyond its 191
attribute and spatial reasoning questions (27 OCR and 14 GPT4V-hard
questions), and 159 questions from HRScene's
Video\_Monitoring subset~\cite{hrscene}.
The latter include all 30 validation questions, all 38 testmini
questions, and 91 questions randomly sampled from the test split. On this held-out
set, we run a coarse-to-fine coordinate grid search
on overall accuracy.
The selected configuration is frozen and applied
unchanged to V$^*$Bench, HRBench-4K, HRBench-8K, and
the general benchmarks in Sec.~\ref{sec:exp_general},
so no evaluation question enters the search.
Table~\ref{tab:hyperparameters} lists the final settings.

All backbones use
four candidate levels, three windows, a single image
presentation, and 3,072 visual tokens per crop.
The preview budgets are 4,096 and 8,192 tokens for the
4B and 8B backbones, respectively; the shared-subspace
projection coefficient and crop context factor are
both fixed at $1$.
We search $d\in\{17,18,19,20,21\}$,
$\lambda\in[0.5,4]$ with a step size of $0.5$, and
$\rho\in[0.1,1]$ with an initial step size of $0.1$,
locally refined to $0.025$.

Answer-readout parameters are searched offline using
cached native logits for all option permutations under
the three input conditions, without additional model
inference. Both $a_v$ and $a_t$ are selected from
\[
\begin{aligned}
\mathcal{A}
&= \{\pm x \mid x\in\mathcal{P}\},\\
\mathcal{P}
&= \{0.3,0.375,0.75,1.5,3,6,\\
&\qquad 8,10,12,24,30,48\}.
\end{aligned}
\]
We initially search $\beta\in[0,1]$ and
$\gamma\in[-1,1]$ with a step size of $0.1$, and
$T_v,T_t$ over $\{0.01,0.1,0.2,\ldots,1\}$.
Their step sizes are successively refined to $0.01$
and $0.001$ when refinement is triggered.

Parameters are updated in a fixed order while the
others remain at the incumbent configuration.
Each sweep retains the two best distinct configurations
and refines, for each refinable parameter, either the
interval between adjacent values or the neighborhood
of a shared value.
Search terminates when no further refinement is
triggered and a complete sweep leaves the incumbent
unchanged.

\begin{table*}[t]
\centering
\caption{Hyperparameters of R$^{3}$Sieve for each backbone.}
\label{tab:hyperparameters}
\setlength{\tabcolsep}{2.5pt}
\renewcommand{\arraystretch}{0.95}
\setlength{\aboverulesep}{0.2ex}
\setlength{\belowrulesep}{0.3ex}
\begin{tabularx}{\textwidth}{@{}cl*{4}{>{\centering\arraybackslash}X}@{}}
\toprule
Module & Hyperparameter & Qwen3-VL-4B & Qwen3-VL-8B & InternVL3.5-4B & InternVL3.5-8B\\
\midrule
\multirow{2}{*}{SSC}
& Response layer $d$ & 18 & 20 & 18 & 18\\
& Preview tokens & 4096 & 8192 & 4096 & 8192\\
\midrule
\multirow{2}{*}{UCS}
& Nonvisual weight $\lambda$ & 1 & 2 & 1.5 & 3\\
& Detail-gain exponent $\rho$ & 1 & 1 & 0.375 & 0.325\\
\midrule
\multirow{6}{*}{CAV}
& Visual pooling exponent $a_{\mathrm v}$ & 0.375 & 3 & 24 & 8\\
& Text pooling exponent $a_{\mathrm t}$ & $-0.375$ & 0.3 & 3 & 0.3\\
& Image-free weight $\beta$ & 0.25 & 0.08 & 0.12 & 0.163\\
& Metadata weight $\gamma$ & 0.22 & $-0.05$ & $-0.20$ & $-0.56$\\
& Visual-branch temperature $T_{\mathrm v}$ & 0.5 & 0.01 & 0.10 & 1\\
& Text-branch temperature $T_{\mathrm t}$ & 1 & 0.25 & 1 & 1\\
\bottomrule
\end{tabularx}
\end{table*}

\subsection{Evaluation Protocol}
\label{app:evaluation_protocol}

\paragraph{Scoring.}
V$^*$Bench has 191 questions, 115 on attributes and 76 on spatial relations. Each HRBench split has 200 semantic questions with 800 official option orders. Predictions are mapped to the official orders for accuracy, and statistical comparisons use semantic questions as the unit. Model weights use BF16, response-scoring matrices use FP64, and target boxes serve only for localization evaluation.

\paragraph{Timing.}
Efficiency uses all questions of each benchmark on one NVIDIA A100-SXM4-80GB GPU, with Qwen3-VL-4B as the answer model. All methods process the same questions with one official option order per question, and R$^{3}$Sieve runs once per semantic input. Table~\ref{tab:efficiency} reports means and sample standard deviations.


\section{Additional Ablation Studies}
\label{app:additional_ablations}

\subsection{Scoring and Window Selection}
\label{app:scoring_selection}

We separate the two choices within UCS using Qwen3-VL-4B
on V$^*$Bench. We cross residual energy
$d_j=\lVert r_j\rVert_2^2$ and leverage scoring
$\phi_\lambda(j)$ with independent ranking and GCS.
Independent ranking selects the three windows with the
largest standalone scores $\eta(u)\sum_{j\in u}w_j$,
where $w_j$ is the chosen location score, without
updating these scores after selection. GCS instead
selects windows by their marginal coverage gain.
All four variants share the SSC residual, candidate
windows, three-crop budget, visual-token budgets,
detail-gain exponent, prompt templates, and CAV
configuration.

\begin{table}[t]
    \centering
    \small
    \caption{Scoring and selection ablation on
    V$^*$Bench (\%).}
    \setlength{\tabcolsep}{7pt}
    \begin{tabular}{@{}lcc@{}}
        \toprule
        Location score & Independent ranking & GCS \\
        \midrule
        Residual energy & 92.67 & 93.72 \\
        Leverage score & 90.05 & \textbf{95.29} \\
        \bottomrule
    \end{tabular}
    \label{tab:scoring_selection}
\end{table}

Table~\ref{tab:scoring_selection} shows that the
advantage of leverage scoring depends on the selection
rule. Under independent ranking, replacing residual
energy with leverage scores reduces accuracy.
With GCS, this ordering reverses, and the gain from
GCS is larger under leverage scoring. Thus, the
improvement of UCS comes from the combination of
scoring and selection. Leverage scores discount
responses repeated across locations, while GCS
discounts coverage already supplied by selected
windows. The observed dependence is consistent with
these two forms of redundancy being addressed jointly.
The paired examples in Appendix~\ref{app:gcs_examples}
further illustrate how GCS changes the selected evidence.

\subsection{CAV Diagnostics}
\label{app:cav_diagnostics}

We further examine CAV with SSC, leverage scoring,
GCS, the selected crops, and the prompts fixed.
Each variant inherits the full configuration except
for the interventions listed in
Table~\ref{tab:cav_diagnostics}, with no retuning.
Setting $\gamma=0$ removes the metadata correction
term while retaining the metadata in the visual input.
Setting $a_v=a_t=1$ makes each pooled branch score
the logarithm of the arithmetic mean of its per-order
probabilities. D2 retains both image-free reference
branches, while D3 retains only the question reference.
D4 sets both branch temperatures to one.
D5 retains pooling without calibration.
D6 uses the original option order without pooling.

\begin{table*}[t]
    \centering
    \small
    \caption{CAV diagnostics on V$^*$Bench (\%).}
    \label{tab:cav_diagnostics}
    \setlength{\tabcolsep}{3pt}
    \renewcommand{\arraystretch}{1.08}
    \begin{tabularx}{\textwidth}{
        @{}l>{\raggedright\arraybackslash}Xccccccc@{}
    }
        \toprule
        ID & Variant & $(a_v,a_t)$
        & $\beta$ & $\gamma$ & $(T_v,T_t)$
        & Attr. & Spa. & All \\
        \midrule
        D0 & Full CAV
        & $(.375,-.375)$ & $.25$ & $.22$ & $(.5,1)$
        & 98.26 & \textbf{90.79} & \textbf{95.29} \\

        D1 & Without metadata correction
        & $(.375,-.375)$ & $.25$ & $0$ & $(.5,1)$
        & 98.26 & 86.84 & 93.72 \\

        D2 & Probability mean with both references
        & $(1,1)$ & $.25$ & $.22$ & $(.5,1)$
        & 98.26 & 88.16 & 94.24 \\

        D3 & Probability mean with question reference
        & $(1,1)$ & $.25$ & $0$ & $(.5,1)$
        & \textbf{99.13} & 85.53 & 93.72 \\

        D4 & Unit branch temperatures
        & $(.375,-.375)$ & $.25$ & $.22$ & $(1,1)$
        & 98.26 & 86.84 & 93.72 \\

        D5 & Pooling without calibration
        & $(.375,-.375)$ & $0$ & $0$ & $(.5,1)$
        & \textbf{99.13} & 85.53 & 93.72 \\

        D6 & Original-order readout
        & -- & $0$ & $0$ & --
        & 98.26 & 85.53 & 93.19 \\
        \bottomrule
    \end{tabularx}
\end{table*}

The metadata correction improves overall accuracy
under both pooling choices, but its gain is smaller
under probability averaging. Its contribution
therefore depends on the pooling rule.
Full CAV also outperforms probability averaging
with the same reference terms and the variant with
unit branch temperatures. These comparisons support
the conditional contributions of the selected
pooling exponents, metadata correction, and temperature
scaling under a frozen configuration.
Some simplified variants achieve higher attribute
accuracy but lower spatial and overall accuracy,
placing the main benefit of full CAV in spatial
judgments. D5 and D6 provide reference readouts,
while component contributions are assessed using
pairs that retain the other settings.

\subsection{BVS and CAV on HRBench}
\label{app_bvs_cav_cases}

BVS uses an oracle that emphasizes the presence of nouns in the
question. This criterion can locate a queried object without ensuring
that the selected evidence resolves the requested attribute or relation.
HRBench includes attribute and counting questions that require more
than object presence. The cases in Table~\ref{tab_bvs_cav_cases}
illustrate this distinction through camera color, sign direction, and
label meaning, each of which requires finer evidence than recognizing
the corresponding entity.

\begin{table}[t]
    \centering
    \small
    \caption{Three paired HRBench cases from a frozen
    Qwen3-VL-4B evaluation with BVS and CAV.}
    \setlength{\tabcolsep}{3pt}
    \renewcommand{\arraystretch}{1.12}
    \begin{tabularx}{\columnwidth}{
        @{}>{\raggedright\arraybackslash}Xrrr@{}
    }
        \toprule
        \multirow{2}{*}{Case}
        & \multicolumn{2}{c}{Visual margin}
        & \multirow{2}{*}{
            \shortstack{Calib.\\advantage}
        } \\
        \cmidrule(lr){2-3}
        & 4K & 8K & \\
        \midrule
        Camera color
        & 0.800 & 10.195 & 0.974 \\
        Zaragoza sign direction
        & 0.198 & 2.490 & 1.019 \\
        Circled label meaning
        & 2.073 & 4.976 & 2.816 \\
        \bottomrule
    \end{tabularx}
    
    \label{tab_bvs_cav_cases}
\end{table}

For each pair, the visual margin compares the correct answer with the
incorrect answer selected by CAV in the 4K instance. We use the same
incorrect alternative for the 8K comparison. The calibration advantage
is the combined contribution of the question and metadata reference
terms in favor of that alternative. These image-free terms are identical
within each pair.

In these cases, the correct answer leads in the pooled visual branch
at both resolutions. The calibration advantage exceeds this lead in
4K and reverses the prediction. The paired 8K instances retain larger
visual margins and remain correct under the same correction. The
relative effect of calibration therefore depends on the visual margin
provided by the selected evidence. These observations support our
hypothesis that BVS evidence can leave a narrow margin for fine-grained
judgments in 4K, making otherwise correct visual predictions more
vulnerable to CAV calibration.

\subsection{Hyperparameter Sensitivity}
\label{app:sensitivity}

\begin{figure*}[t]
    \centering
    \includegraphics[width=\textwidth]{
        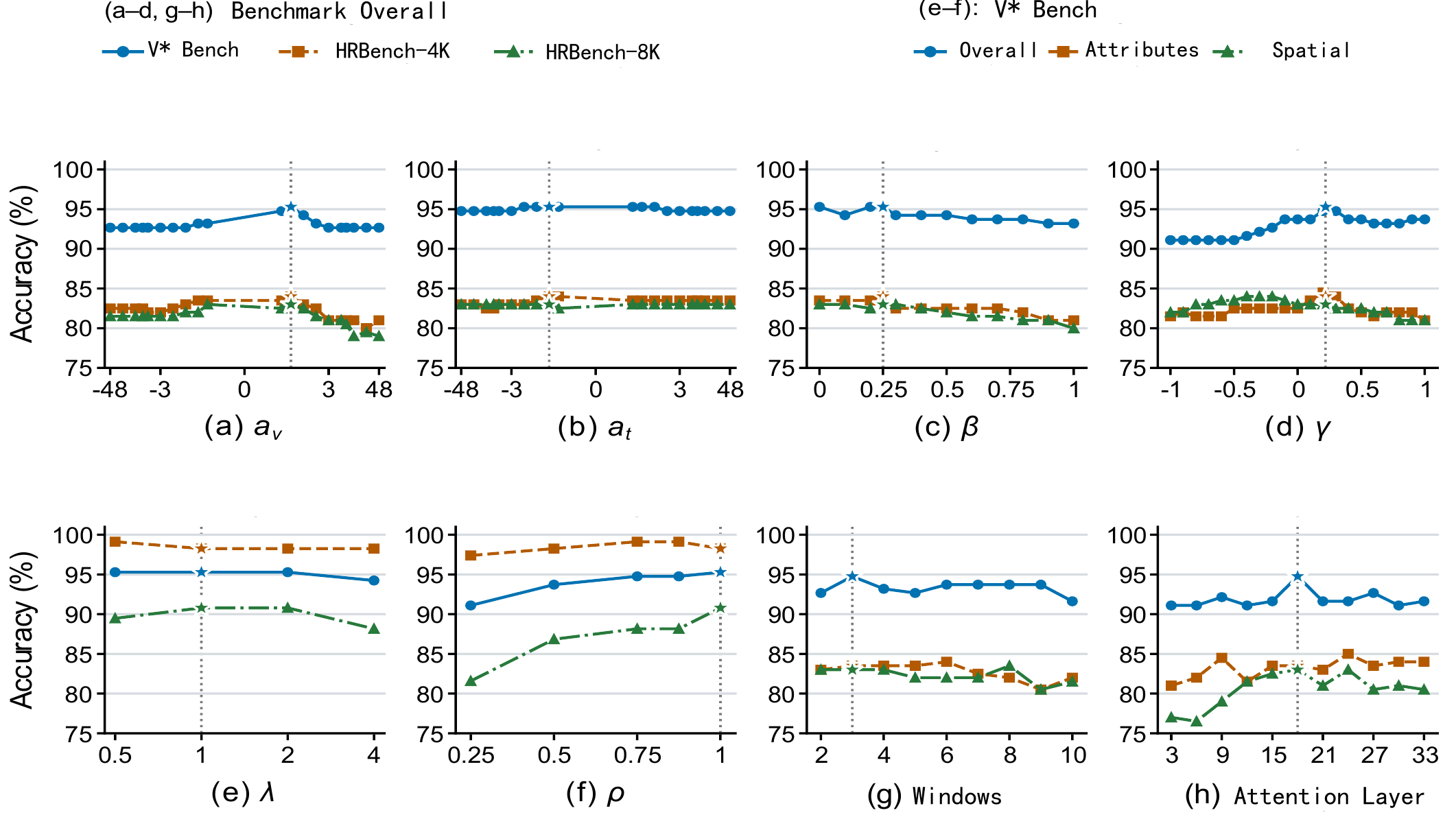
    }
    \caption{Sensitivity analysis of different hyperparameters.}
    \label{fig:exp_sensitivity}
\end{figure*}

We analyze Qwen3-VL-4B by varying one hyperparameter at a
time (Fig.~\ref{fig:exp_sensitivity}). Overall, performance is relatively stable across the tested values of $a_v$, $a_t$, $\beta$, $\gamma$, and $\lambda$. Increasing the detail-gain exponent $\rho$ mainly improves spatial accuracy, while attribute accuracy changes comparatively little. The preferred attention layer varies across datasets, and three windows provide the best practical configuration.

\section{Complementary Window Examples}
\label{app:gcs_examples}

\begin{figure*}[t]
    \centering
    \includegraphics[width=0.98\textwidth]{
        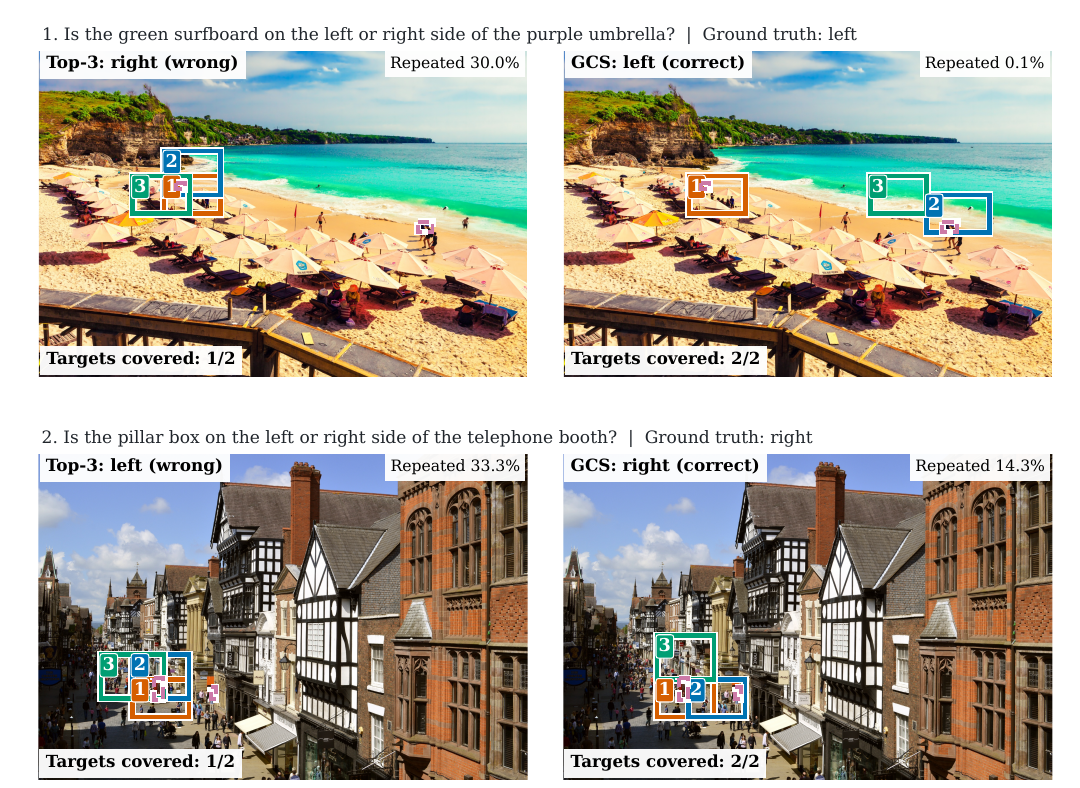
    }
    \caption{Independent ranking (left) and GCS (right) on two spatial-relation questions.}
    \label{fig:gcs_window_ablation}
\end{figure*}

Figure~\ref{fig:gcs_window_ablation} compares independent top-three ranking (left) with GCS (right) on two spatial-relation questions. GCS discounts locations that the selected windows already cover and thereby reaches both queried entities. Orange, blue, and green outlines give the selection order, and dashed magenta outlines mark the target boxes.

\section{Attention Map Visualizations}
\label{app:attention_visualizations}

\begin{figure*}[t]
    \centering
    \includegraphics[
        width=\textwidth,
        height=0.84\textheight,
        keepaspectratio
    ]{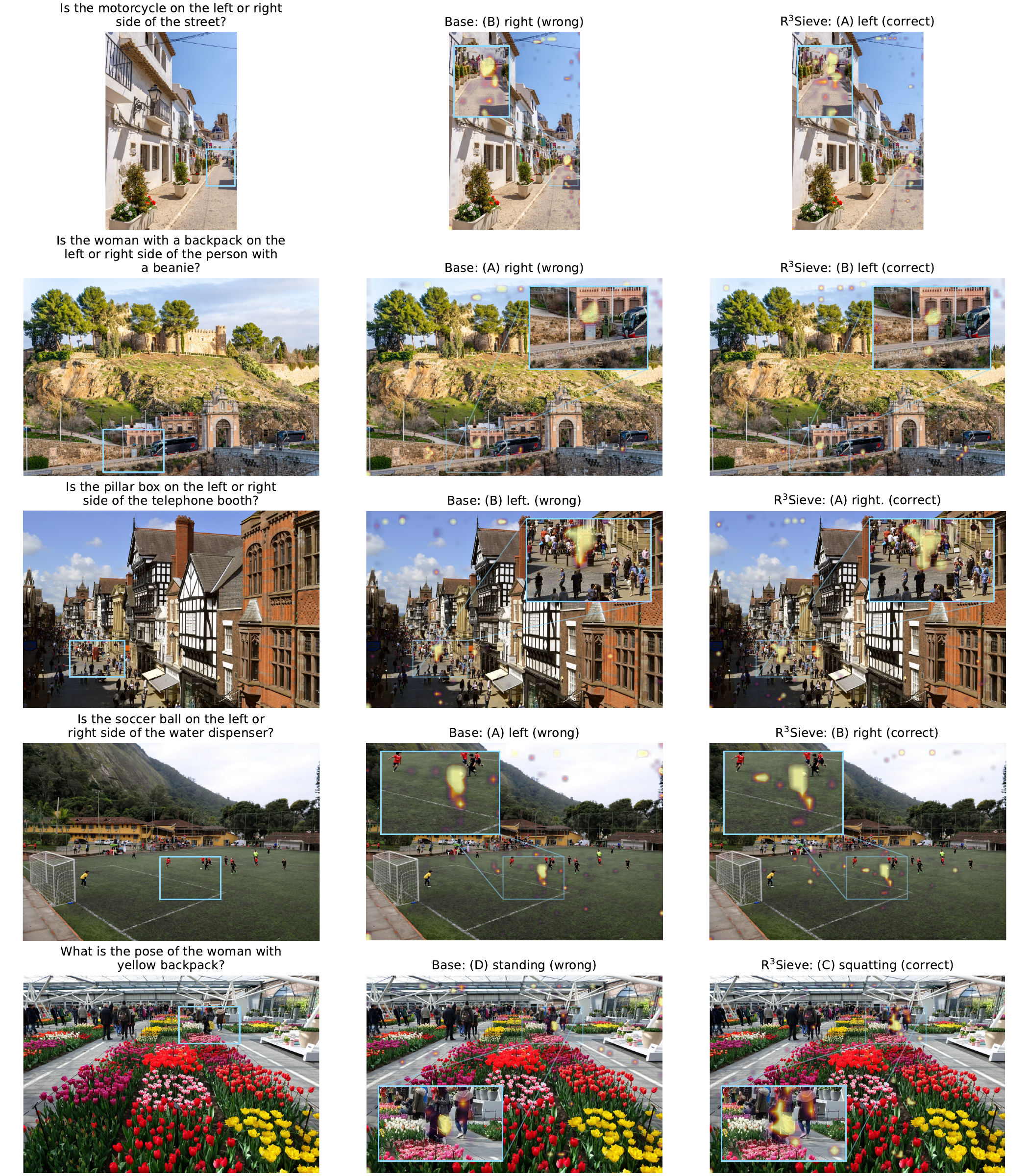}
    \caption{Base and R$^{3}$Sieve layer-18 maps on five V$^*$Bench examples.}
    \label{fig:response_attention_gallery}
\end{figure*}

\begin{figure*}[p]
    \centering
    \includegraphics[
        width=\textwidth,
        height=0.84\textheight,
        keepaspectratio
    ]{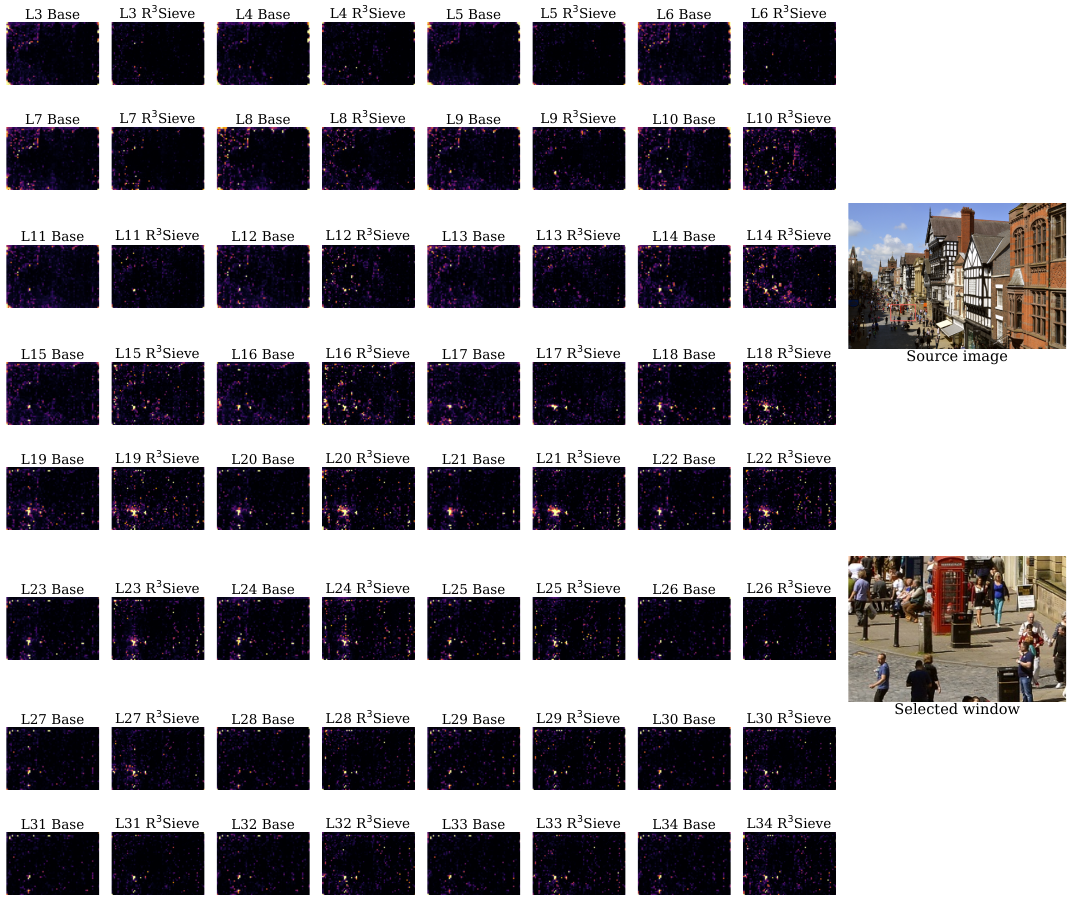}
    \caption{Base and R$^{3}$Sieve maps across layers 3--34 for the telephone-booth example.}
    \label{fig:telephone_attention}
\end{figure*}

Figure~\ref{fig:response_attention_gallery} pairs the layer-18 responses of Base and R$^{3}$Sieve with their selected windows and answers on five V$^*$Bench cases. R$^{3}$Sieve places more of the visible response on the queried entities and selects crops that keep the evidence for the attribute or spatial judgment, and it answers all five questions correctly, while Base does not. Figure~\ref{fig:telephone_attention} traces the telephone-booth case through layers 3 to 34, where shared-subspace cancellation separates the question-relevant region from diffuse scene responses and layer 18 provides the map used for selection. Each map has its own color scale, so only the spatial focus is comparable across maps.

\end{document}